\PassOptionsToPackage{hyphens}{url}

\newcommand{\underreview}{0}
\newcommand{\mergewithsuppement}{1}

\documentclass[acmtog]{acmart}

\acmSubmissionID{411}
\usepackage{graphicx}
\usepackage{ifthen}
\usepackage{hyperref}
\usepackage{float}
\usepackage{morefloats}
\usepackage{alltt}
\usepackage{amsmath}
 
\usepackage{amssymb}
\usepackage{amsthm}
\usepackage{ulem}
\usepackage{mathrsfs}
\usepackage{newlfont} %
\usepackage{ulem}
\usepackage{floatflt}
\usepackage{wrapfig}
\usepackage[ruled]{algorithm2e} %
\usepackage{booktabs} %
\usepackage{gensymb}
\usepackage{fixltx2e}
\usepackage{subfig} %
\usepackage{multirow}
\usepackage{cleveref}
\usepackage[T1]{fontenc} %
\usepackage{etoolbox}
\usepackage{xfrac}
\usepackage{booktabs} %
\usepackage[export]{adjustbox}
\usepackage{algpseudocode}
\usepackage[acronym,nomain]{glossaries} 
\usepackage{todonotes}
\usepackage{adjustbox}
\usepackage{xcolor}
\usepackage{wrapfig}

\newcommand{\etal}{{et~al.~}}

\newcommand{\nothing}[1]{}

\definecolor{AudioColor}{rgb}{0.56,0.34,0.62}

\definecolor{VideoColor}{rgb}{0.44,0.66,0.38}

\definecolor{DeadlineColor}{rgb}{0.9,0.4,0}

\definecolor{OldColor}{rgb}{0.5,0.5,0.5}

\definecolor{NewColor}{rgb}{0.9,0.4,0}

\definecolor{DeleteColor}{rgb}{0.1,0.6,1.0}
\newcommand{\delete}[1]{{\color{DeleteColor} #1}}
\definecolor{MoveColor}{rgb}{0.5,0.1,0.5}

\definecolor{figred}{rgb}{1,0,0}
\definecolor{figgreen}{rgb}{0,0.6,0}
\definecolor{figblue}{rgb}{0,0,1}
\definecolor{figpink}{rgb}{1,0.63,0.63}

\newcommand{\pseudocode}{Pseudocode}
\floatstyle{plain}
\newfloat{algorithm}{tbhp}{lop}
\floatname{algorithm}{\pseudocode}

\ifthenelse{\isundefined{\diff}}
{

\renewcommand{\delete}[1]{}

}
{
\ifthenelse{\equal{\diff}{0}}
{

\renewcommand{\delete}[1]{}

}
{}
}

\ifdefined\google

\else

\fi

\newcommand{\filename}[1]{\url{#1}}
\newcommand{\foldername}[1]{\url{#1}}

\catcode`_=12
\begingroup\lccode`~=`_\lowercase{\endgroup\let~\sb}
\mathcode`_="8000

\newcommand{\colvector}[1]{\textbf{#1}}
\def\eigenval{\lambda}
\def\covariance{\Omega}
\def\anisotropy{A}
\def\denoise{D}
\def\robustness{R}
\def\robustnessfinal{\hat{\robustness}}
\def\kernel{k}
\def\kernelbase{\kernel_{detail}}
\def\kerneldenoise{\kernel_{denoise}}
\def\structure{\widehat{\Omega}}
\def\kernelstretch{\kernel_{stretch}}
\def\kernelshrink{\kernel_{shrink}}

\def\denoisethreshold{\denoise_{th}}
\def\denoisetransition{\denoise_{tr}}

\def\tilesize{T_{s}}

\def\rejectionstrength{s}
\def\rejectionthreshold{t}

\def\displacement{d}

\def\mv{v}
\def\motionstr{M}
\def\motionstrthreshold{\motionstr_{th}}

\def\neighborhood{N}
\newcommand{\neighbor}[1]{\neighborhood_#1}

\def\measured{ms}
\def\modelled{md}

\def\robustnesssigma{\sigma}
\def\robustnesssigmameasured{\robustnesssigma_{\measured}}
\def\robustnesssigmamodelled{\robustnesssigma_{\modelled}}

\def\robustnesscolordiff{d}
\def\robustnesscolordiffmeasured{\robustnesscolordiff_{\measured}}
\def\robustnesscolordiffmodelled{\robustnesscolordiff_{\modelled}}

\acmPrice{15.00}
\setcopyright{rightsretained}
\acmJournal{TOG}
\acmYear{2019}\acmVolume{38}\acmNumber{4}\acmArticle{28}\acmMonth{7} \acmDOI{10.1145/3306346.3323024}

\setcitestyle{square}

\begin{document}
\normalem

\title[Handheld Multi-Frame Super-Resolution]{Handheld Multi-Frame Super-Resolution}
\ifthenelse{\equal{\underreview}{1}}
{
  \renewcommand{\shortauthors}{}
}
{
\author{Bartlomiej Wronski} \email{bwronski@google.com}
\author{Ignacio Garcia-Dorado} \email{ignaciod@google.com}
\author{Manfred Ernst} \email{ernstm@google.com}
\author{Damien Kelly} \email{damienkelly@google.com}
\author{Michael Krainin} \email{mkrainin@google.com}
\author{Chia-Kai Liang} \email{ckliang@google.com}
\author{Marc Levoy} \email{levoy@google.com}
\author{Peyman Milanfar} \email{milanfar@google.com}
\affiliation{ Google Research\streetaddress{1600 Amphitheatre Parkway}\city{Mountain View}\state{CA}\postcode{94043}}

\renewcommand{\shortauthors}{Wronski \etal}
}

\begin{abstract}
Compared to DSLR cameras, smartphone cameras have smaller sensors, which limits their spatial resolution; smaller apertures, which limits their light gathering ability; and smaller pixels, which reduces their signal-to-noise ratio.
The use of color filter arrays (CFAs) requires demosaicing, which further degrades resolution.
In this paper, we supplant the use of traditional demosaicing in single-frame and burst photography pipelines with a multi-frame super-resolution algorithm that creates a complete RGB image directly from a burst of CFA raw images.
We harness natural hand tremor, typical in handheld photography, to acquire a burst of raw frames with small offsets.
These frames are then aligned and merged to form a single image with red, green, and blue values at every pixel site.
This approach, which includes no explicit demosaicing step, serves to both increase image resolution and boost signal to noise ratio.
Our algorithm is robust to challenging scene conditions: local motion, occlusion, or scene changes.
It runs at 100 milliseconds per 12-megapixel RAW input burst frame on mass-produced mobile phones.
Specifically, the algorithm is the basis of the \textit{Super-Res Zoom} feature, as well as the default merge method in \textit{Night Sight} mode (whether zooming or not) on Google's flagship phone.
\end{abstract}




\begin{teaserfigure}
  \centering
  \includegraphics[width=\linewidth]{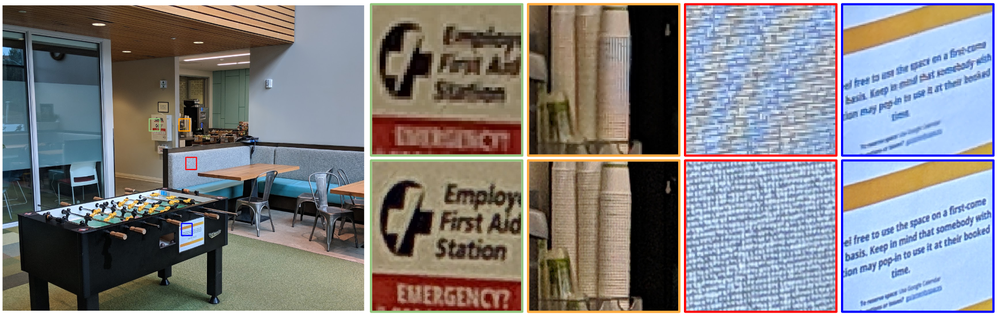}
  \vspace{-0.2in}
  \caption{We present a multi-frame super-resolution algorithm that supplants the need for demosaicing in a camera pipeline by merging a burst of raw images.
  We show a comparison to a method that merges frames containing the same-color channels together first, and is then followed by demosaicing (\textbf{top}).
  By contrast, our method (\textbf{bottom}) creates the full RGB directly from a burst of raw images.
  This burst was captured with a hand-held mobile phone and processed on device.
  Note in the third (red) inset that the demosaiced result exhibits aliasing (Moir\'{e}), while our result takes advantage of this aliasing, which changes on every frame in the burst, to produce a merged result in which the aliasing is gone but the cloth texture becomes visible.}
  \label{fig:teaser}
\end{teaserfigure}

\maketitle

\section{Introduction}\label{sec:intro}
Smartphone camera technology has advanced to the point that taking pictures with a smartphone has become the most popular form of photography~\cite{CIPA:2018:CR,Flickr:2017:Report}.
Smartphone photography offers high portability and convenience, but many challenges still exist in the hardware and software design of a smartphone camera that must be overcome to enable it to compete with dedicated cameras.

Foremost among these challenges is limited spatial resolution. The resolution produced by digital image sensors is limited not only by the physical pixel count (e.g., 12-megapixel camera), but also by the presence of color filter arrays (CFA)\footnote{Also known as a color filter mosaic (CFM).} like the Bayer CFA \cite{Bayer:1976:CIA}.
Given that human vision is more sensitive to green, a quad of pixels in the sensor usually follows the Bayer pattern RGGB; i.e., 50\% green, 25\% red, and 25\% blue.
The final full-color image is generated from the spatially undersampled color channels through an interpolation process called demosaicing~\cite{Li:2008:ID}.

Demosaicing algorithms operate on the assumption that the color of an area in a given image is relatively constant.
Under this assumption, the color channels are highly correlated, and the aim of demosaicing is to reconstruct the undersampled color information while avoiding the introduction of any visual artifacts.
Typical artifacts of demosaicing include false color artifacts such as chromatic aliases, zippering (abrupt or unnatural changes of intensity over consecutive pixels that look like a zipper), maze, false gradient, and Moir\'{e} patterns (\Cref{fig:teaser} top).
Often, the challenge in effective demosaicing is trading off resolution and detail recovery against introducing visual artifacts.
In some cases, the underlying assumption of cross-channel correlation is violated, resulting in reduced resolution and loss of details.

A significant advancement in smartphone camera technology in recent years has been the application of software-based computational photography techniques to overcome limitations in camera hardware design.
Examples include techniques for increasing dynamic range~\cite{Hasinoff:2016:BPH}, improving signal-to-noise ratio through denoising~\cite{Mildenhall:2018:BDK,Godard:2018:DBD} and wide aperture effects to synthesize shallow depth-of-field~\cite{Wadhwa:2018:SDF}.
Many of these recent advancements have been achieved through the introduction of \emph{burst processing}\footnote{We use the terms multi-frame and \emph{burst processing} interchangeably to refer to the process of generating a single image from multiple images captured in rapid succession.} where on a shutter press multiple acquired images are combined to produce a photo that is of greater quality than that of a single acquired image.

In this paper, we introduce an algorithm that uses signals captured across multiple shifted frames to produce higher resolution images (\Cref{fig:teaser} bottom).
Although the underlying techniques can be generalized to any shifted signals, in this work we focus on applying the algorithm to the task of resolution enhancement and denoising in a smartphone image acquisition pipeline using burst processing.
By using a multi-frame pipeline and combining different undersampled and shifted information present in different frames, we remove the need for an explicit demosaicing step.

To work on a smartphone camera, any such algorithm must:
\begin{itemize}
\item \textbf{Work handheld from a single shutter press}
  -- without a tripod or deliberate motion of the camera by the user.
\item \textbf{Run at an interactive rate}
  -- the algorithm should produce the final enhanced resolution with low latency (within at most a few seconds).
\item \textbf{Be robust to local motion and scene changes}
  -- users might capture scenes with fast moving objects or scene changes.
  While the algorithm might not increase resolution in all such scenarios, it should not produce appreciable artifacts.
\item \textbf{Be robust to noisy input data}
  -- in low light the algorithm should not amplify noise, and should strive to reduce it.
\end{itemize}

With these criteria in mind, we have developed an algorithm that processes multiple successively captured raw frames in an online fashion.
The algorithm tackles the tasks of demosaicing and super-resolution jointly and formulates the problem as the reconstruction and interpolation of a continuous signal from a set of sparse samples.
Red, green and blue pixels are treated as separate signals on different planes and reconstructed simultaneously.
This approach enables the production of highly detailed images even when there is no cross-channel correlation -- as in the case of saturated single-channel colors.
The algorithm requires no special capturing conditions; natural hand-motion produces offsets that are sufficiently random in the subpixel domain to apply multi-frame super-resolution.
Additionally, since our super-resolution approach creates a continuous representation of the input, it allows us to directly create an image with a desired target magnification / zoom factor without the need for additional resampling.
The algorithm works on a mobile device and incurs a computational cost of only 100 ms per 12-megapixel processed frame.

The main contributions of this work are:
\begin{enumerate}
  \item Replacing raw image demosaicing with a multi-frame super-resolution algorithm.
  \item The introduction of an adaptive kernel interpolation / merge method from sparse samples (\Cref{sec:multiframe}) that takes into account the local structure of the image, and adapts accordingly.
  \item A motion robustness model (\Cref{subsec:robustness}) that allows the algorithm to work with bursts containing local motion, disocclusions, and alignment/registration failures (\Cref{fig:motion_prior}).
  \item The analysis of natural hand tremor as the source of subpixel coverage sufficient for super-resolution (\Cref{sec:hand_motion}).
\end{enumerate}

\begin{figure*}[ht]
  \centering
  \includegraphics[width=\linewidth]{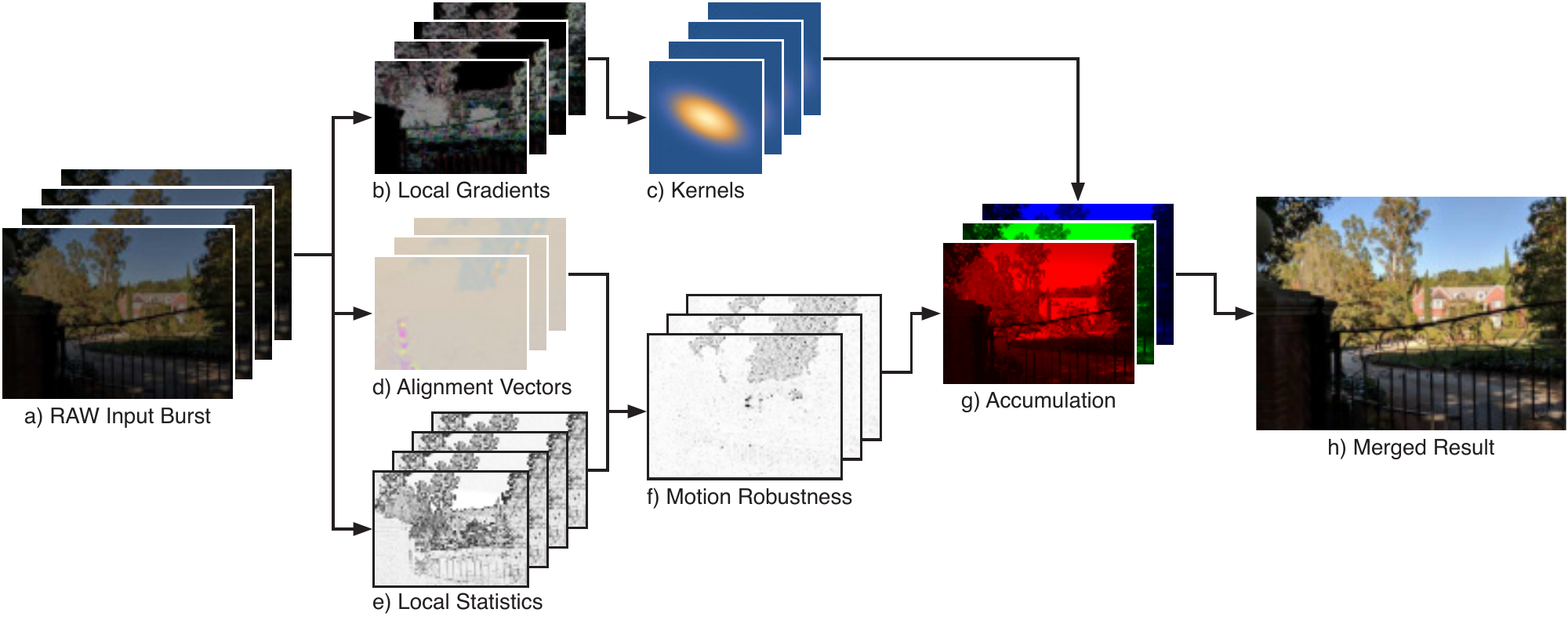}
  \vspace{-0.3in}
  \caption{\textbf{Overview of our method:}
  A captured burst of raw (Bayer CFA) images \textbf{(a)} is the input to our algorithm.
  Every frame is aligned locally \textbf{(d)} to a single frame, called the \emph{base frame}.
  We estimate each frame's contribution at every pixel through kernel regression (\Cref{subsec:kernel_merge}).
  These contributions are accumulated separately per color channel \textbf{(g)}.
  The kernel shapes \textbf{(c)} are adjusted based on the estimated local gradients \textbf{(b)} and the sample contributions are weighted based on a robustness model \textbf{(f)} (\Cref{subsec:robustness}).
  This robustness model computes a per-pixel weight for every frame using the alignment field \textbf{(d)} and local statistics \textbf{(e)} gathered from the neighborhood around each pixel.
  The final merged RGB image \textbf{(h)} is obtained by normalizing the accumulated results per channel.
  We call the steps depicted in \textbf{(b)}-\textbf{(g)} the \emph{merge}.}
  \label{fig:overview}
\end{figure*}

\section{Background}\label{sec:related_work}

\subsection{Demosaicing}
Demosaicing has been studied extensively~\cite{Li:2008:ID}, and the literature presents a wide range of algorithms.
Most methods interpolate the missing green pixels first (since they have double sampling density) and reconstruct the red and blue pixel values using color ratio~\cite{Lukac:2004:NCR} or color difference~\cite{Hirakawa:2006:JDD}.
Other approaches work in the frequency domain~\cite{Leung:2011:LSL}, residual space~\cite{Monno:2015:ARI}, use LAB homogeneity metrics~\cite{Hirakawa:2005:AHD} or non local approaches~\cite{Duran:2014:SSS}.
More recent works use CNNs to solve the demosaicing problem such as the joint demosaicing and denoising technique by Gharbi \etal ~\shortcite{Gharbi:2016:DJD}.
Their key insight is to create a better training set by defining metrics and techniques for mining difficult patches from community photographs.

\subsection{Multi-frame Super-resolution (SR)}
Single-image approaches exploit strong priors or training data.
They can \emph{suppress} aliasing\footnote{In this work we refer to aliasing in signal processing terms -- a signal with frequency content above half of the sampling rate that manifests as a lower frequency after sampling~\cite{Nyquist:1928:CTT}.} well, but are often limited in how much they can \textit{reconstruct} from aliasing.
In contrast to single frame techniques, the goal of multi-frame super-resolution is to increase the true (optical) resolution.

In the sampling theory literature, multi-frame super-resolution techniques date as far back as the '50s~\cite{Yen:1956:ONS} and the '70s~\cite{Papoulis:1977:GSE}.
The work of Tsai~\shortcite{Tsai:1984:MIR} started the modern concept of super-resolution by showing that it was possible to improve resolution by registering and fusing multiple aliased images.
Irani and Peleg~\shortcite{Irani:1991:IRI}, and then Elad and Feuer~\shortcite{Elad:1997:RSS} formulated the algorithmic side of super-resolution.
The need for accurate subpixel registration, existence of aliasing, and good signal-to-noise levels were identified as the main requirements of practical super-resolution~\cite{Baker:2012:LSR,Robinson:2004:FPL,Robinson:2006:SPA}.

In the early 2000s, Farsiu \etal ~\shortcite{Farsiu:2006:MDS} and Gotoh and Okutomi~\shortcite{Gotoh:2004:DSR} formulated super-resolution from arbitrary motion as an optimization problem that would be infeasible for interactive rates.
Ben-Ezra \etal ~\shortcite{Ben:2005:VSR} created a jitter camera prototype to do super-resolution using controlled subpixel detector shifts.
This and other works inspired some commercial cameras (e.g.,~\textit{Sony A6000}, \textit{Pentax FF K1}, \textit{Olympus OM-D E-M1} or \textit{Panasonic Lumix DC-G9}) to adopt multi-frame techniques, using controlled pixel shifting of the physical sensor.
However, these approaches require the use of a tripod or a static scene.
Video super-resolution approaches~\cite{Belekos:2010:MAPVSR,Liu:2011:BAAVSR,Sajjadi:2018:FRVSR} counter those limitations and extend the idea of multi-frame super-resolution to video sequences.

\subsection{Kernel Based Super-resolution and Interpolation}
Takeda \etal ~\shortcite{Takeda:2006:RKR,Takeda:2006:KRI} formulated super-resolution as a kernel regression and reconstruction problem, which allows for faster processing.
Around the same time, M{\"u}ller \etal ~\shortcite{Muller:2005:PFI} introduced a technique to model fluid-fluid interactions that can be rendered using kernel methods introduced by Blinn~\shortcite{Blinn:1982:AGA}.
Yu and Turk~\shortcite{Yu:2013:RSP} proposed an adaptive solution to the reconstructing of surfaces of particle-based fluids using anisotropic kernels.
These kernels, like Takeda \etal's, are based on local gradient Principal Component Analysis (PCA), where the anisotropy of the kernels allows for simultaneous preservation of sharp features and smooth rendering of flat surfaces.
Similar adaptive kernel based method were proposed for single image super-resolution by Hunt~\shortcite{Hunt:2004:ISR} and for general upscaling and interpolation by Lee and Yoon~\shortcite{Lee:2010:NIU}.
We adopt some of these ideas and generalize them to fit our use case.

\subsection{Burst Photography and Raw Fusion}
Burst fusion methods based on raw imagery are relatively uncommon in the literature, as they require knowledge of the photographic pipeline~\cite{Farsiu:2006:MDS,Heide:2014:FAF,Wu:2006:TCV,Gotoh:2004:DSR}.
Vandewalle \etal ~\shortcite{Vandewalle:2007:JDS} described an algorithm where information from multiple Bayer frames is separated into luminance and chrominance components and fused together to improve the CFA demosaicing.
Most relevant to our work is Hasinoff \etal ~\shortcite{Hasinoff:2016:BPH} which introduced an end-to-end burst photography pipeline fusing multiple frames for increased dynamic range and signal-to-noise ratio.
Our paper is a more general fusion approach that (a) dispenses with demosaicing, (b) produces increased resolution, and (c) enables merging onto an arbitrary grid, allowing for high quality digital zoom at modest factors (\Cref{sec:discussion}).
Most recently, Li \etal ~\shortcite{Li:2018:DPI} proposed an optimization based algorithm for forming an RGB image directly from fused, unregistered raw frames.

\subsection{Multi-frame Rendering}
This work also draws on multi-frame and temporal super-resolution techniques widely used in real-time rendering (for example, in video games).
Herzog \etal combined information from multiple rendered frames to increase resolution \shortcite{Herzog:2010:STU}.
Sousa \etal ~\shortcite{Jimenez:2011:FAR} mentioned the first commercial use of robustly combining information from two frames in real-time in a video game, while Malan~\shortcite{Malan:2012:RTG} expanded its use to produce a $1920 \times 1080$ image from four $1280 \times 720$ frames.
Subsequent work~\cite{Sousa:2013:GGC,Drobot:2014:HRA,Karis:2014:HQT} established temporal super-resolution techniques as state-of-the-art and standard in real-time rendering for various effects, including dynamic resolution rendering and temporal denoising.
Salvi~\shortcite{Salvi:2016:AET} provided a theoretical explanation of commonly used local color neighborhood clipping techniques and proposed an alternative based on statistical analysis.
While most of those ideas are used in a different context, we generalize their insights about detecting aliasing, misalignment and occlusion in our robustness model.

\subsection{Natural Hand Tremor}
While holding a camera (or any object), a natural, involuntary hand tremor is always present.
The tremor is comprised of low amplitude and high frequency motion components that occur while holding steady limb postures~\cite{Schafer:1886:RMR}.
The movement is highly periodic, with a frequency in the range of 8--12~Hz, and its movement is small in magnitude but random~\cite{Marshall:1956:PT}.
The motion also consists of a \emph{mechanical-reflex} component that depends on the limb, and a second component that causes micro-contractions in the limb muscles~\cite{Riviere:1998:ACP}.
This behavior has been shown to not change with age~\cite{Sturman:2005:EAR} but it can change due to disease~\cite{NIH:2018:TF}.
In this paper, we show that the hand tremor of a user holding a mobile camera is sufficient to provide subpixel coverage for super-resolution.

\section{Overview of Our Method}\label{sec:overview}
Our approach is visualized in \Cref{fig:overview}.
First, a burst of raw (CFA Bayer) images is captured.
For every captured frame, we align it locally with a single frame from the burst (called the \emph{base frame}).
Next, we estimate each frame's local contributions through kernel regression (\Cref{subsec:kernel_merge}) and accumulate those contributions across the entire burst.
The contributions are accumulated separately per color plane.
We adjust kernel shapes based on the estimated signal features and weight the sample contributions based on a robustness model (\Cref{subsec:robustness}).
We perform per-channel normalization to obtain the final merged RGB image.

\subsection{Frame Acquisition}
Since our algorithm is designed to work within a typical \emph{burst processing} pipeline, it is important that the processing does not increase the overall photo capture latency.
Typically, a smartphone operates in a mode called \emph{Zero-Shutter Lag}, where raw frames are being captured continuously to a ring buffer when the user opens and operates the camera application.
On a shutter press, the most recent captured frames are sent to the camera processing pipeline.
Our algorithm operates on an input burst (\Cref{fig:overview} (a)) formed from those images.
Relying on previously captured frames creates challenges for the super-resolution algorithm -- the user can be freely moving the camera prior to the capture.
The merge process must be able to deal with natural hand motion (\Cref{sec:hand_motion}) and cannot require additional movement or user actions.

\subsection{Frame Registration and Alignment}\label{subsec:frame_registration} %
Prior to combining frames, we place them into a common coordinate system by registering frames against the \emph{base frame} to create a set of alignment vectors (\Cref{fig:overview} (d)).
Our alignment solution is a refined version of the algorithm used by Hasinoff \etal~\shortcite{Hasinoff:2016:BPH}.
The core alignment algorithm is coarse-to-fine, pyramid-based block matching that creates a pyramid representation of every input frame and performs a limited window search to find the most similar tile.
Through the alignment process we obtain per patch/tile (with tile sizes of $\tilesize$) alignment vectors relative to the base frame.

Unlike Hasinoff \etal~\shortcite{Hasinoff:2016:BPH}, we require subpixel accurate alignment to achieve super-resolution.
To address this issue we could use a different, dedicated registration algorithm designed for accurate subpixel estimation (e.g., Fleet and Jepson~\shortcite{Fleet:1990:CCI}), or refine the block matching results.
We opted for the latter due to its simplicity and computational efficency.
We have explored estimating the subpixel offsets by fitting a quadratic curve to the block matching alignment error and finding its minimum~\cite{Kanade:1991:SMA}; however, we found that super-resolution requires a more accurate method.
Therefore, we refine the block matching alignment vectors by three iterations of Lucas-Kanade~\shortcite{Lucas:1981:IIR} optical flow image warping.
This approach reached the necessary accuracy while keeping the computational cost low.

\subsection{Merge Process}
After frames are aligned, the remainder of the merge process (\Cref{fig:overview} (b-g)) is responsible for fusing the raw frames into a full RGB image.
These steps constitute the core of our algorithm and will be described in greater detail in \Cref{sec:multiframe}.

The merge algorithm works in an online fashion, sequentially computing the contributions of each processed frame to every output pixel by accumulating colors from a $3\times3$ neighborhood.
Those contributions are weighted by kernel weights (\Cref{subsec:kernel_merge}, \Cref{fig:overview} (c)), modulated by the robustness mask (\Cref{subsec:robustness}, \Cref{fig:overview} (f)), and summed together separately for the red, green and blue color planes.
At the end of the process, we divide the accumulated color contributions by the accumulated weights, obtaining three color planes.

The result of the merge process is a full RGB image, which can be defined at any desired resolution.
This can be processed further by the typical camera pipeline (spatial denoising, color correction, tone-mapping, sharpening) or alternatively saved for further offline processing in a non-CFA raw format like \textit{Linear DNG}~\cite{Adobe:2012:DNG}.

Before we explain the algorithm details, we analyze the key characteristics that enable the hand-held super-resolution.

\begin{figure}[t!]
  \centering
  \includegraphics[width=\linewidth]{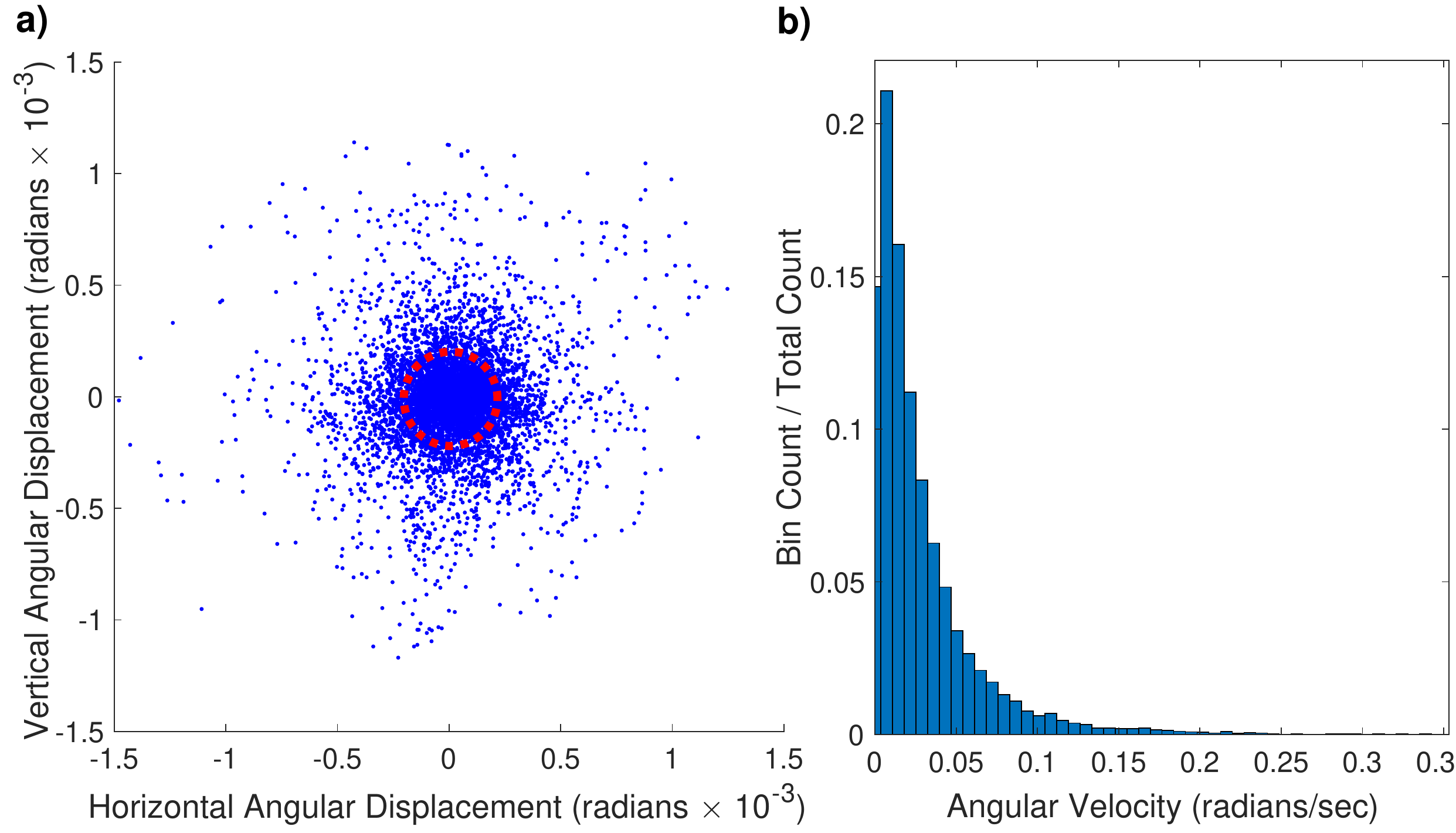}
  \caption{\textbf{(a)} Horizontal and vertical angular displacement (i.e., \textbf{not} including translational displacement) arising from handheld motion evaluated over a test set of $86$ captured bursts. The red circle corresponds to one standard deviation (which maps to a pixel displacement magnitude of \textbf{0.89 pixels}) showing that the distribution is roughly symmetrical. \textbf{(b)} Histogram of angular velocity magnitude measured over the test set showing that during captures the rotational velocity remains relatively low.}
  \label{fig:handheld}
\end{figure}

\begin{figure*}[t!]
  \captionsetup{farskip=0pt}
  \centering
  \includegraphics[width=\textwidth]{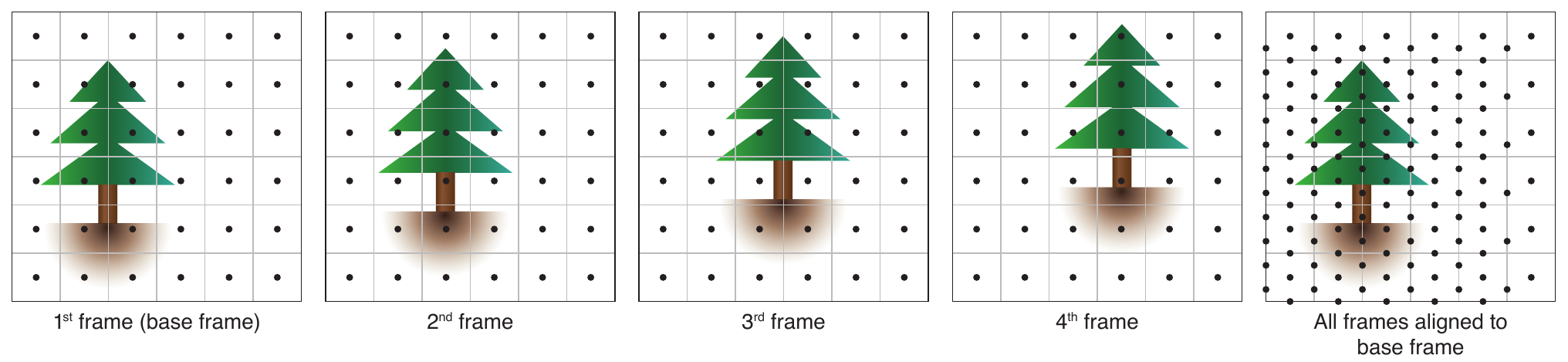}
  \vspace{-0.2in}
  \caption{\textbf{Subpixel displacements from handheld motion:}
    Illustration of a burst of four frames with linear hand motion.
    Each frame is offset from the previous frame by half a pixel along the x-axis and a quarter pixel along the y-axis due to the hand motion.
    After alignment to the base frame, the pixel centers (black dots) uniformly cover the resampling grid (grey lines) at an increased density.
    In practice, the distribution is more random than in this simplified example.
  }
  \vspace{-0.15in}
  \label{fig:subpixel_distribution}
\end{figure*}

\section{Hand-held Super-Resolution}\label{sec:hand_motion}
Multi-frame super-resolution requires two conditions to be fulfilled~\cite{Tsai:1984:MIR}:
\begin{enumerate}
  \item Input frames need to be aliased, i.e., contain high frequencies that manifest themselves as false low frequencies after sampling.
  \item The input must contain multiple aliased images, sampled at different subpixel offsets.
  This will manifest as different phases of false low frequencies in the input frames.
\end{enumerate}
Having multiple lower resolution shifted and aliased images allows us to both remove the effects of aliasing in low frequencies as well as reconstruct the high frequencies.
In a (mobile) camera pipeline, $(1)$ means an image sensor having distances between pixels larger than the spot size of the lens.
Our algorithm assumes that the input raw frames are aliased (see discussion in \Cref{sec:discussion}).

Most existing multi-frame super-resolution approaches impose restrictions on the types of motion noted in $(2)$.
This includes commercially available products like DSLR cameras using sensor shifts of a camera placed on a tripod.
Those requirements are impractical for casual photography and violate our algorithm goals; therefore we make use of natural hand motion.
Some publications like Li \etal ~\shortcite{Li:2018:DPI} use unregistered, randomly offset images for the purpose of super-resolution or multi-frame demosaicing, however to our knowledge no prior work analyzes if the subpixel coverage produced by hand tremor is sufficient to obtain consistent results.
We show that using hand tremor alone is enough to move the device adequately during the burst acquisition.

To analyze the actual behavior when capturing bursts of photographs while holding a mobile device, we have examined the hand movement in a set of $86$ bursts.
The analyzed bursts were captured by $10$ different users during casual photography and not for the purpose of this experiment.
Mobile devices provide precise information about rotational movement measured by a gyroscope, which we use in our analysis.
As we lack measurements about translation of the device, we ignore translations in this analysis, although we recognize that they also occur.
In \Cref{subsec:robustness}, we show how our algorithm is robust to parallax, and occlusions or disocclusions, caused by translational camera displacement.

First, we used the phone gyroscope rotational velocity measurements and integrated them to find the relative rotation of the phone compared to the burst capture start.
We plotted them along with a histogram of angular velocities in \Cref{fig:handheld}.
Our analysis confirms that the hand movement introduces uniformly random (no directions are preferred) angular displacements and relatively slow rotation of the capture device during the burst acquisition.
The following section analyzes movement in the subpixel space and how it facilitates random sampling.

\subsection{Handheld Motion in subpixel Space}\label{subsec:hand_motion_sup_pixel}
Although handshake averaged over the course of a long time interval is random and isotropic, handshake over the course of a short burst might be nearly a straight line or gentle curve in X-Y \cite{Hee:2014:GYR}.
Will this provide a uniform enough distribution of subpixel samples?
It does, but for non-obvious reasons.

Consider each pixel as a point sample, and assume a pessimistic, least random scenario -- that the hand motion is regular and linear.
After alignment to the base frame, the point samples from all frames combined will be approximately uniformly distributed in the subpixel space (Figure~\ref{fig:subpixel_distribution}).
This follows from the equidistribution theorem~\cite{Weyl:1910:GE}, which states that the sequence $\{a, 2a, 3a, \ldots \mod 1\}$ is uniformly distributed (if $a$ is an irrational number).
Note that while the equidistribution theorem assumes infinite sequences, the closely related concept of rank-1 lattices is used in practice to generate finite point sets with low discrepancy for image synthesis~\cite{Dammertz:2008:R1L} in computer graphics.

\begin{figure}[t]
  \captionsetup{farskip=0pt}
  \centering
  \includegraphics[width=\linewidth]{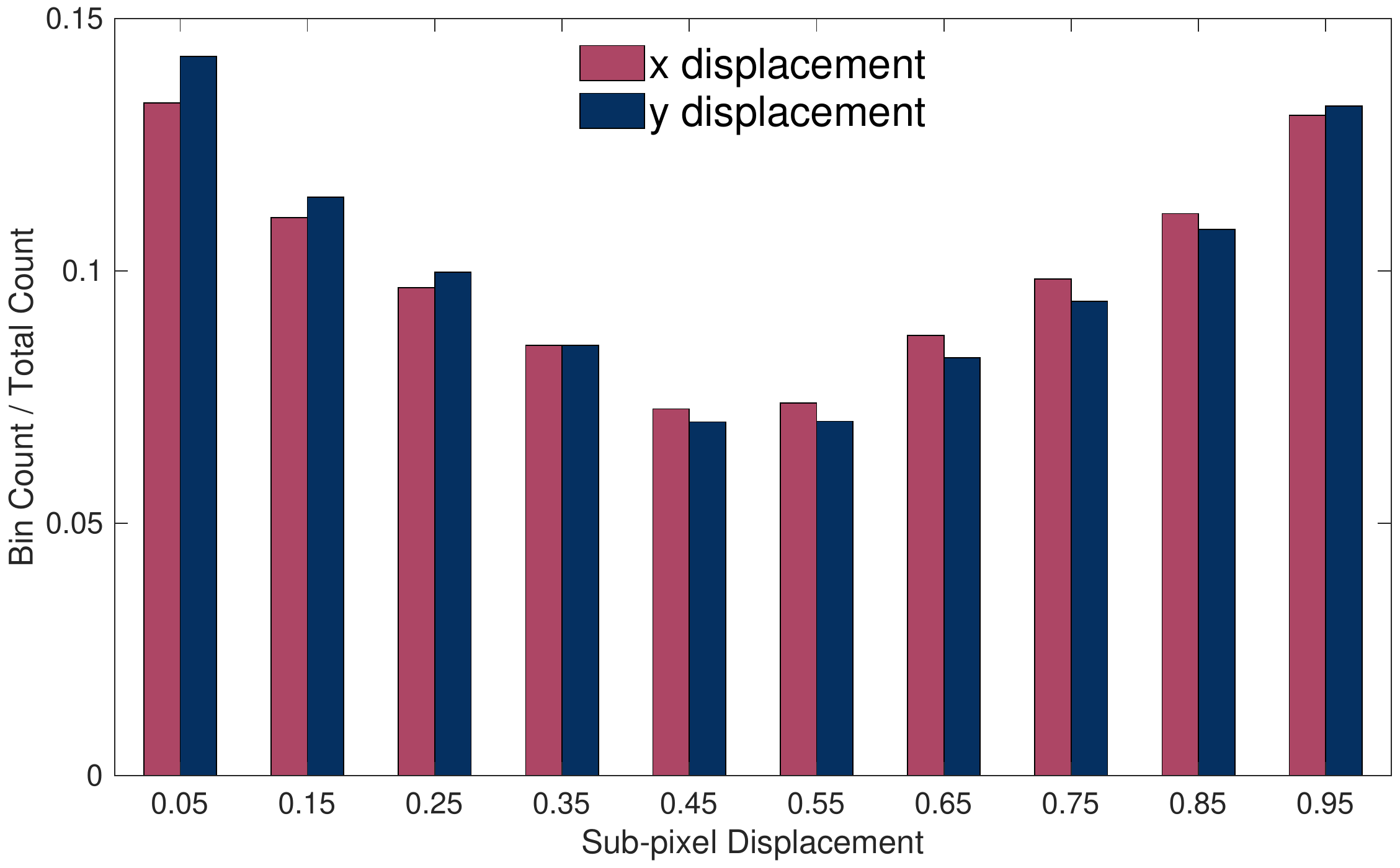}
  \vspace{-0.25in}
  \caption{\textbf{Distribution of estimated subpixel displacements:}
    Histogram of x and y subpixel displacements as computed by the alignment algorithm (\Cref{subsec:frame_registration}).
    While the alignment process is biased towards whole-pixel values, we observe sufficient coverage of subpixel values to motivate super-resolution.
    Note that displacements in x and y are not correlated.}
  \vspace{-0.2in}
  \label{fig:handheld_subpixel}
\end{figure}

Obviously, not all the assumptions above hold in practice.
Therefore, we verified empirically that the resulting sample locations are indeed distributed as expected.
We measured the subpixel offsets by registration (\Cref{subsec:frame_registration}) for $16 \times 16$ tiles aggregated across $20$ handheld burst sequences.
The biggest deviation from a uniform distribution is caused by a phenomenon known as pixel locking and is visible in the histogram as bias towards whole pixel values.
As can be seen in \Cref{fig:handheld_subpixel}, pixel locking causes non-uniformity in the distribution of subpixel displacements.
Pixel locking is an artifact of any subpixel registration process~\cite{Robinson:2004:FPL, Shimiziu:2005:SPE} that depends on the image content (high spatial frequencies and more aliasing cause a stronger bias).
Despite this effect, the subpixel coverage of displacements remains sufficiently large in the range $(0, 1)$ to motivate the application of super-resolution.

\section{Proposed Multi-Frame Super-Resolution Approach}\label{sec:multiframe}
Super-resolution techniques reconstruct a high resolution signal from multiple lower resolution representations.
Given the stochastic nature of pixel shifts resulting from natural hand motion, a good reconstruction technique to use in our case is kernel regression (\Cref{subsec:kernel_merge}) that reconstructs a continuous signal.
Such continuous representation can be resampled at any resolution equal to or higher than the original input frame resolution (see \Cref{sec:discussion} for discussion of the effective resolution).
We use anisotropic Gaussian Radial Basis Function (RBF) kernels (\Cref{subsec:kernels}) that allow for locally adaptive detail enhancement or spatio-temporal denoising.
Finally, we present a robustness model (\Cref{subsec:robustness}) that allows our algorithm to work in scenes with complex motion and to degrade gracefully to single frame upsampling in cases where alignment fails.

\subsection{Kernel Reconstruction}\label{subsec:kernel_merge}
The core of our algorithm is built on the idea of treating pixels of multiple raw Bayer frames as irregularly offset, aliased and noisy measurements of three different underlying continuous signals, one for each color channel of the Bayer mosaic.
Though the color channels are often correlated, in the case of saturated colors (for example red, green or blue only) they are not.
Given sufficient spatial coverage, separate per-channel reconstruction allows us to recover the original high resolution signal even in those cases.

\begin{figure}[t!]
  \centering
  \includegraphics[width=\linewidth]{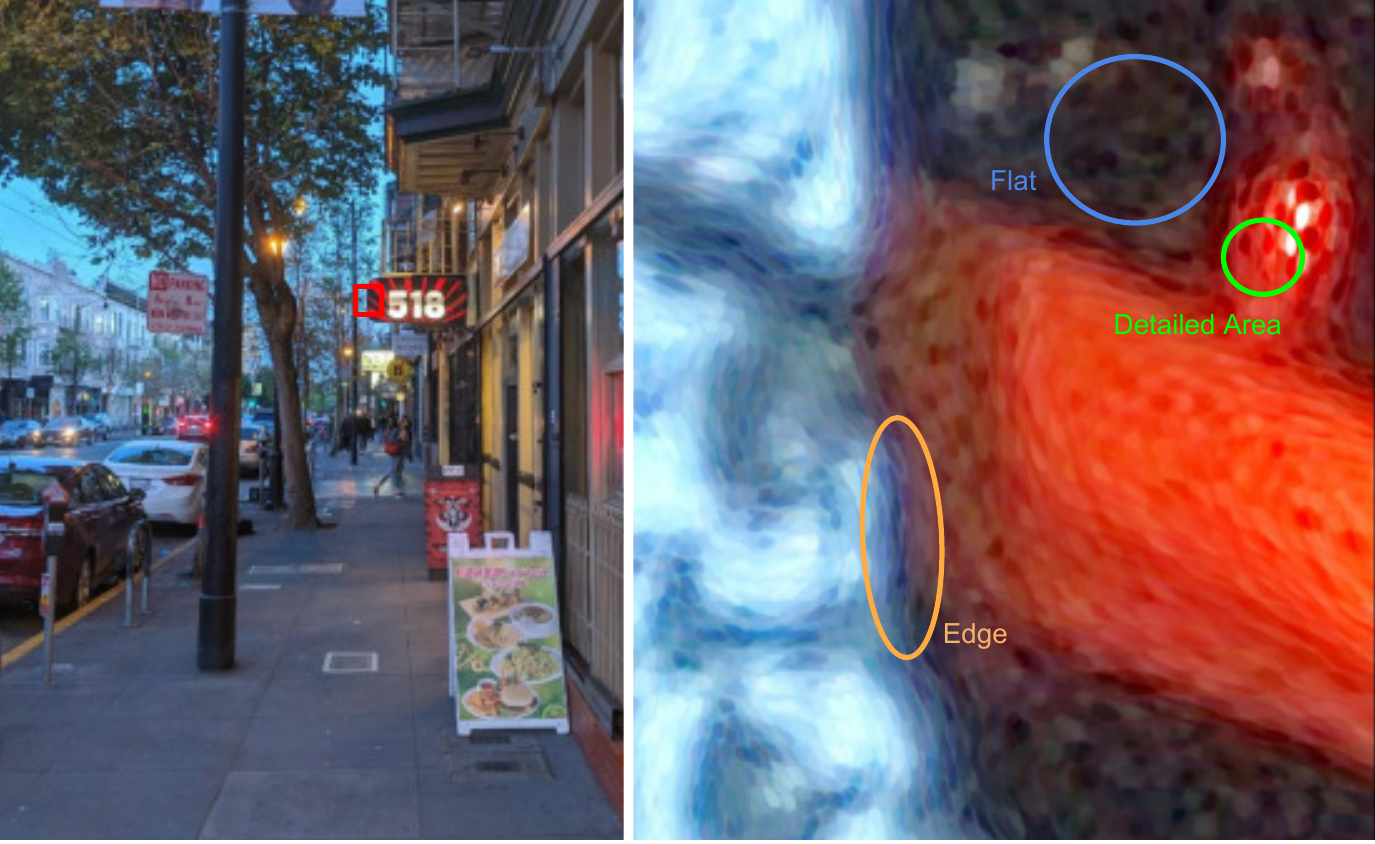}
  \caption{\textbf{Sparse data reconstruction with anisotropic kernels:}
  Exaggerated example of very sharp (i.e., narrow, $\kernelbase=0.05px$) kernels on a real captured burst. For demonstration purposes, we represent samples corresponding to whole RGB input pictures instead of separate color channels.
  Kernel adaptation allows us to apply differently shaped kernels on edges (orange), flat (blue) or detailed areas (green).
  The orange kernel is aligned with the edge, the blue one covers a large area as the region is flat, and the green one is small to enhance the resolution in the presence of details.}
  \label{fig:sparse_reconstruction_2d}
\end{figure}

To produce the final output image we processes all frames sequentially -- for every output image pixel, we evaluate local contributions to the red, green and blue color channels from different input frames.
Every input raw image pixel has a different color channel, and it contributes only to a specific output color channel.
Local contributions are weighted; therefore, we accumulate weighted contributions and weights. At the end of the pipeline, those contributions are normalized.
For each color channel, this can be formulated as:
\begin{equation}
C(x,y)=\frac{\sum_{n}\sum_{i} c_{n,i} \cdot w_{n,i} \cdot \robustnessfinal_n}{\sum_{n}\sum_{i} w_{n,i} \cdot \robustnessfinal_n},
\end{equation}
where $(x,y)$ are the pixel coordinates, the sum $\sum_{n}$ is over all contributing frames, $\sum_{i}$ is a sum over samples within a local neighborhood (in our case 3$\times$3), $c_{n,i}$ denotes the value of the Bayer pixel at given frame $n$ and sample $i$, $w_{n,i}$ is the local sample weight and $\robustnessfinal_n$ is the local robustness (\Cref{subsec:robustness}).
In the case of the \emph{base frame}, $\robustnessfinal$ is equal to $1$ as it does not get aligned, and we have full confidence in its local sample values.

To compute the local pixel weights, we use local radial basis function kernels, similarly to the non-parametric kernel regression framework of Takeda \etal ~\shortcite{Takeda:2006:RKR,Takeda:2006:KRI}.
Unlike Takeda et~al., we don't determine kernel basis function parameters at sparse sample positions.
Instead, we evaluate them at the final resampling grid positions.
Furthermore, we always look at the nine closest samples in a $3 \times 3$ neighborhood and use the same kernel function for all those samples.
This allows for efficient parallel evaluation on a GPU.
Using this "gather" approach every output pixel is independently processed only once per frame.
This is similar to work of Yu and Turk~\shortcite{Yu:2013:RSP}, developed for fluid rendering.
Two steps described in the following sections are: estimation of the kernel shape (\Cref{subsec:kernels}) and robustness based sample contribution weighting (\Cref{subsec:robustness}).

\begin{figure}[t!]
  \centering
  \includegraphics[width=\linewidth]{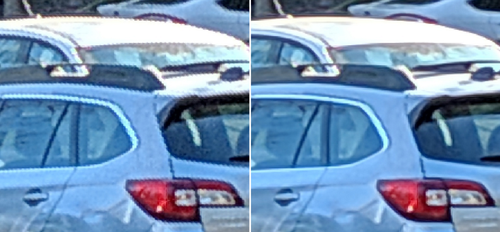}
  \caption{\textbf{Anisotropic Kernels:}
    \textbf{Left:} When isotropic kernels ($\kernelstretch = 1$, $\kernelshrink = 1$, see supplemental material) are used, small misalignments cause heavy zipper artifacts along edges.
    \textbf{Right:} Anisotropic kernels ($\kernelstretch = 4$, $\kernelshrink = 2$) fix the artifacts.}
  \label{fig:fig_anisotropic_result}
\end{figure}

\subsubsection{Local Anisotropic Merge Kernels}\label{subsec:kernels}

Given our problem formulation, kernel weights and kernel functions define the image quality of the final merged image: kernels with wide spatial support produce noise-free and artifact-free, but blurry images, while kernels with very narrow support can produce sharp and detailed images.
A natural choice for kernels used for signal reconstruction are \textit{Radial Basis Function} kernels - in our case anisotropic Gaussian kernels.
We can adjust the kernel shape to different local properties of the input frames: amounts of detail and the presence of edges (\Cref{fig:sparse_reconstruction_2d}).
This is similar to kernel selection techniques used in other sparse data reconstruction applications \cite{Takeda:2006:RKR,Takeda:2006:KRI,Yu:2013:RSP}.

Specifically, we use a 2D unnormalized anisotropic Gaussian RBF for $w_{n,i}$:
\begin{equation}
w_{n,i}=\exp\left(-\frac{1}{2}\displacement_i^T \covariance^{-1} \displacement_i\right),
\end{equation}
where $\covariance$ is the kernel covariance matrix and $\displacement_i$ is the offset vector of sample $i$ to the output pixel ($\displacement_i = \left[x_i - x_0, y_i - y_0\right] ^T$).

One of the main motivations for using anisotropic kernels is that they increase the algorithm's tolerance for small misalignments and uneven coverage around edges.
Edges are ambiguous in the alignment procedure (due to the aperture problem) and result in alignment errors ~\cite{Robinson:2004:FPL} more frequently compared to non-edge regions of the image.
Subpixel misalignment as well as a lack of sufficient sample coverage can manifest as \textit{zipper artifacts} (\Cref{fig:fig_anisotropic_result}).
By stretching the kernels along the edges, we can enforce the assignment of smaller weights to pixels not belonging to edges in the image.

\subsubsection{Kernel Covariance Computation}\label{subsec:kernel_covariance}

\begin{figure}[t!]
  \centering
  \includegraphics[width=0.7\linewidth]{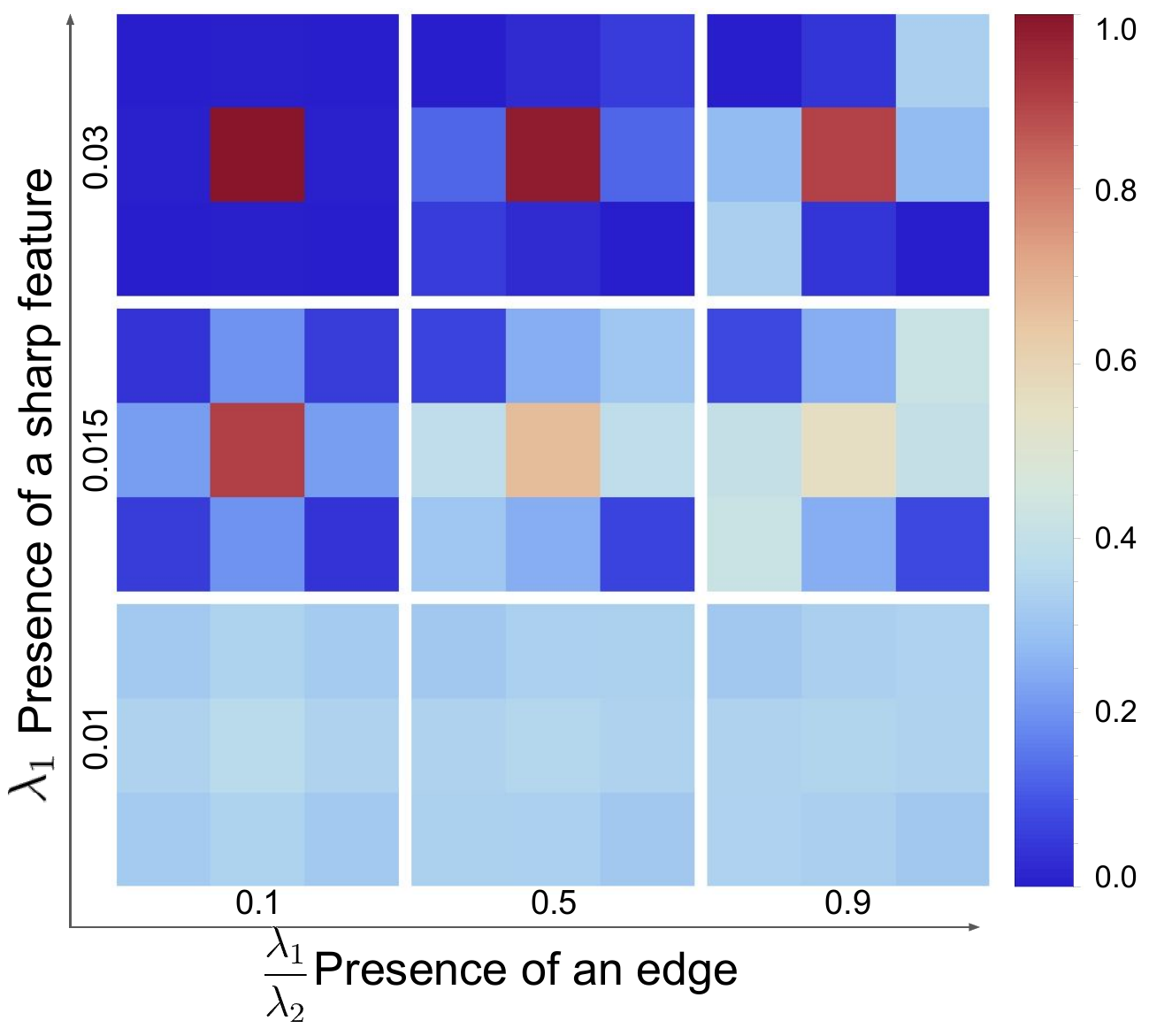}
  \vspace{-0.1in}
  \caption{\textbf{Merge kernels:} Plots of relative weights in different $3\times3$ sampling kernels as a function of local tensor features.}
  \label{fig:kernel_plots}
\end{figure}

We compute the kernel covariance matrix by analyzing every frame's local gradient structure tensor.
To improve runtime performance and resistance to image noise, we analyze gradients of half-resolution images formed by decimating the original raw frames by a factor of two.
To decimate a Bayer image containing different color channels, we create a single pixel from a $2\times2$ Bayer quad by combining four different color channels together.
This way, we can operate on single channel luminance images and perform the computation at a quarter of the full resolution cost and with improved signal-to-noise ratio.
To estimate local information about strength and direction of gradients, we use gradient structure tensor analysis ~\cite{Harris:1988:ACC,Bigun:1991:MOE}:
\begin{equation}
\structure=\begin{bmatrix}
I^2_x     & I_x I_y\\
I_x I_y & I^2_y
\end{bmatrix},
\end{equation}
where $I_x$ and $I_y$ are the local image gradients in horizontal and vertical directions, respectively.
The image gradients are computed by finite forward differencing the luminance in a small, $3\times3$ color window (giving us four different horizontal and vertical gradient values).
Eigenanalysis of the local structure tensor $\structure$ gives two orthogonal direction vectors $\colvector{e}_1$, $\colvector{e}_2$ and two associated eigenvalues $\lambda_1$, $\lambda_2$.
From this, we can construct the kernel covariance as:
\begin{equation}\label{eq:kernel_matrix}
\covariance =
\begin{bmatrix}
\colvector{e}_1 &\colvector{e}_2
\end{bmatrix}
\begin{bmatrix}
\kernel_1 & 0\\
0 & \kernel_2
\end{bmatrix}
\begin{bmatrix}
\colvector{e}^T_1\\
\colvector{e}^T_2
\end{bmatrix},
\end{equation}
where $\kernel_1$ and $\kernel_2$ control the desired kernel variance in either edge or orthogonal direction.
We control those values to achieve adaptive super-resolution and denoising.
We use the magnitude of the structure tensor's dominant eigenvalue $\eigenval_1$ to drive the spatial support of the kernel and the trade-off between the super-resolution and denoising, where $\frac{\eigenval_1-\eigenval_2}{\eigenval_1+\eigenval_2}$ is used to drive the desired anisotropy of the kernels (\Cref{fig:kernel_plots}).
The specific process we use to compute the final kernel covariance can be found in the supplemental material along with the tuning values.
Since $\covariance$ is computed at half of the Bayer image resolution, we upsample the kernel covariance values through bilinear sampling before computing the kernel weights.

\subsection{Motion Robustness}\label{subsec:robustness} %

\begin{figure}[t!]
  \centering
  \includegraphics[width=1.0\linewidth]{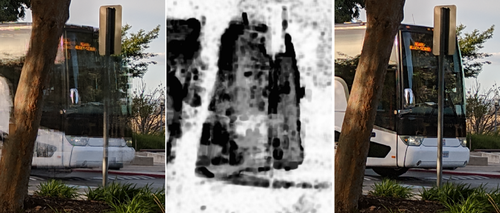}
  \vspace{-0.2in}
  \caption{\textbf{Motion robustness:} \textbf{Left:} Photograph of a moving bus without any robustness model.
    Alignment errors and occlusions correspond to severe tiling and ghosting artifacts.
    \textbf{Middle:} an accumulated robustness mask produced by our model.
    White regions correspond to all frames getting merged and contributing to super-resolution, while dark regions have a smaller number of merged frames because of motion or incorrect alignment.
    \textbf{Right:} result of merging frames with the robustness model.}
  \vspace{-0.15in}
  \label{fig:fig_rejection}
\end{figure}

Reliable alignment of an arbitrary sequence of images is extremely challenging -- because of both theoretical~\cite{Robinson:2004:FPL} and practical (available computational power) limitations.
Even assuming the existence of a perfect registration algorithm, changes in scene and occlusion can result in some areas of the photographed scene being unrepresented in many frames of the sequence.
Without taking this into account, the multi-frame fusion process as described so far would produce strong artifacts.
To fuse any sequence of frames robustly, we assign confidence to the local neighborhood of every pixel that we consider merging.
We call an image map with those confidences a \emph{robustness mask} where a value of one corresponds to fully merged regions and value of zero to rejected areas (\Cref{fig:fig_rejection}).

\subsubsection{Statistical Robustness Model}

\begin{figure}[t!]
  \centering
  \includegraphics[width=1.0\linewidth]{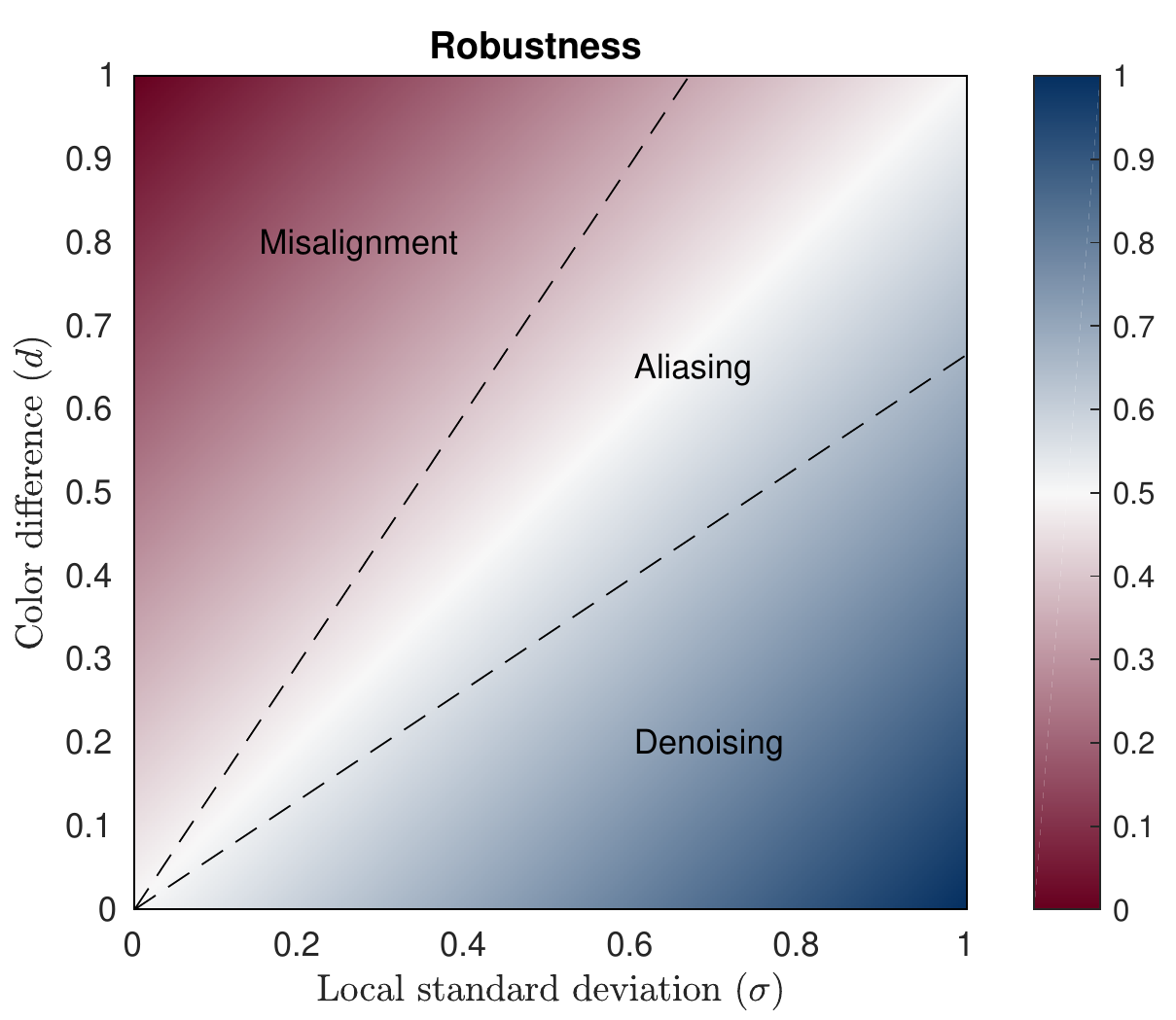}
  \vspace{-0.15in}
  \caption{\textbf{Statistical robustness model:} The relationship between color difference $\robustnesscolordiff$ and local standard deviation $\robustnesssigma$ dictates how we merge a given frame with respect to the base frame.}
  \label{fig:fig_aliasing_aware}
\end{figure}

The core idea behind our robustness logic is to address the following question: how can we distinguish between aliasing, which is necessary for super-resolution, and frame misalignment which hampers it?
We observe that areas prone to aliasing have large spatial variance even within a single frame.
This idea has previously been successfully used in temporal anti-aliasing techniques for real-time graphics ~\cite{Salvi:2016:AET}.
Though our application in fusing information from multiple frames is different, we use a similar local variance computation to find the highly aliased areas.
We compute the local standard deviation in the images $\robustnesssigma$ and a color difference between the base frame and the aligned input frame $\robustnesscolordiff$.
Regions with differences smaller than the local standard deviation are deemed to be non-aliased and are merged, which contributes to temporal denoising.
Differences close to a pre-defined fraction of spatial standard deviation \footnote{that depends on the presence of the motion in the scene, see \Cref{subsec:motion_prior}.} are deemed to be aliased and are also merged, which contributes to super-resolution.
Differences larger than this fraction most likely signify misalignments or non-aligned motion, and are discarded.
Through this analysis, we interpret the difference in terms of standard deviations (\Cref{fig:fig_aliasing_aware}) as the probability of frames being safe to merge using a soft comparison function:
\begin{equation}
\robustness =  \rejectionstrength \cdot \exp\left(-\frac{\robustnesscolordiff^2}{\robustnesssigma^2}\right) - \rejectionthreshold ,
\label{eq:robustness}
\end{equation}
where $\rejectionstrength$ and $\rejectionthreshold$ are tuned scale and threshold parameters used to guarantee that small differences get a weight of one, while large difference get fully rejected.
The following subsections will describe how we compute $\robustnesscolordiff$ and $\robustnesssigma$ as well as how we adjust the $\rejectionstrength$ tuning based on the presence of local motion.

\begin{figure}[t!]
  \centering
  \includegraphics[width=\linewidth]{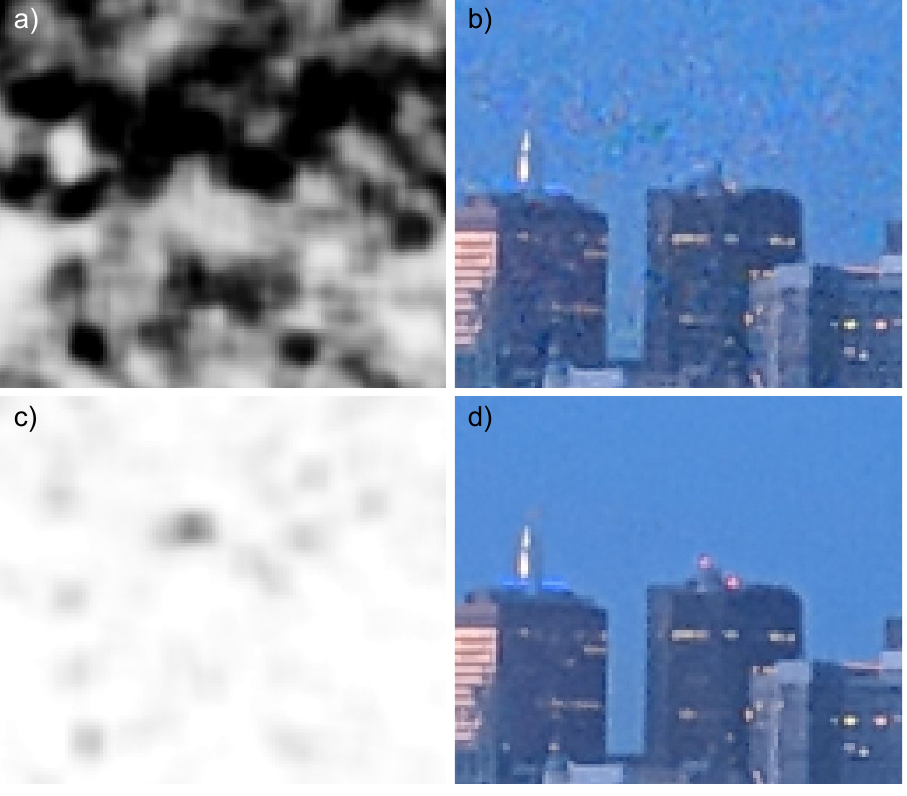}
  \caption{\textbf{Noise model:} \textbf{Top row:} well aligned low-light photo merged without the noise model.
    \textbf{Bottom row:} The same photo merged with the noise model included.
    \textbf{Left:} Accumulated robustness mask.
    \textbf{Right:} The merged image.
    Including the noise model in the statistical comparisons helps to avoid false low confidence in the case of relatively flat, noisy regions.}
  \label{fig:fig_noise_modeling}
\end{figure}

\subsubsection{Noise-corrected Local Statistics and Color Differences}\label{subsec:noisemodel}
First, we create a half-resolution RGB image that we call the \emph{guide image}.
This guide image is formed by creating a single RGB pixel corresponding to each Bayer quad by taking red and blue values directly and averaging the green channels together.
In this section, we will use the following notation: subscript $\measured$ signifies variables measured and computed locally from the guide image, $\modelled$ denotes ones computed according to the noise estimation for given brightness level, and variables without those subscripts are noise-corrected measurements.
For every pixel of the guide image, we compute the color mean and spatial standard deviation $\robustnesssigmameasured$ in a $3 \times 3$ neighborhood.
The local mean is used to compute local color difference $\robustnesscolordiffmeasured$ between the base frame and the aligned input frame.
Since the estimates for $\robustnesssigmameasured$ and $\robustnesscolordiffmeasured$ are produced from a small number of samples, we need to correct them for the expected amount of noise in the image.

Raw images taken with different exposure times and ISOs have different levels of noise.
The noise present in raw images is a heteroscedastic Gaussian noise~\cite{Foi:2008:PPG}, with the noise variance being a linear function of the input signal brightness.
Parameters of this linear function (slope and intercept) depend on the sensor and exposure parameters, which we call the \emph{noise model}.
In low light, noise causes even correctly aligned images to have a much larger $\robustnesscolordiffmeasured$ differences compared to the good lighting scenario.
We estimate the $\robustnesssigmameasured$ and $\robustnesscolordiffmeasured$ from just nine samples for the red and blue color pixels ($3\times3$ neighborhood) and are in effect unreliable due to noise.

To correct those noisy measurements, we incorporate the \emph{noise model} in two ways: we compute the spatial color standard deviation $\robustnesssigmamodelled$ and mean differences between two frames $\robustnesscolordiffmodelled$ that are expected on patches of constant brightness.
We obtain $\robustnesssigmamodelled$ and $\robustnesscolordiffmodelled$ through a series of Monte Carlo simulations for different brightness levels to take into account non-linearities like sensor clipping values around the white point.
Modelled variables are used to clamp $\robustnesssigmameasured$ from below by $\robustnesssigmamodelled$ and to apply a Wiener shrinkage~\cite{Kuan:1985:ANS} on $\robustnesssigmameasured$ to compute final values of $\robustnesssigma$ and $\robustnesscolordiff$:
\begin{equation}
\begin{split}
&\robustnesssigma=\max(\robustnesssigmameasured,\robustnesssigmamodelled).\\
&\robustnesscolordiff=\robustnesscolordiffmeasured\frac{\robustnesscolordiffmeasured^2}{\robustnesscolordiffmeasured^2+\robustnesscolordiffmodelled^2},
\end{split}
\end{equation}
Inclusion of the \emph{noise model} allows us to correctly merge multiple noisy frames in low-light scenario (\Cref{fig:fig_noise_modeling}).

\subsubsection{Additional Robustness Refinement}\label{subsec:motion_prior}

\begin{figure}[t!]
  \centering
  \includegraphics[width=\linewidth]{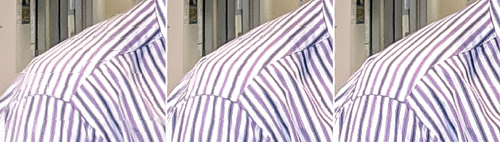}
  \caption{
  \textbf{Left:} Merged photograph of a moving person without any robustness. Misalignment causes fusion artifacts.
  \textbf{Middle:} Merged image with statistical robustness only, some artifacts are still present (we advise the reader to zoom in).
  \textbf{Right:} Final merged image with both statistical robustness and motion prior. Inclusion of both motion robustness terms helps to avoid fusion artifacts.
  }
  \label{fig:motion_prior}
\end{figure}

To improve the robustness further, we use additional information that comes from analyzing local values of the alignment vectors.
We observe that in the case of just camera motion and correct alignment, the alignment field is generally smooth.
Therefore regions with no alignment variation can be attributed to areas with no local motion.
Combining this motion prior into the robustness calculation can remove many more artifacts, as shown in \Cref{fig:motion_prior}.

In the case of misalignments due to the aperture problem or presence of local motion in the scene, the local alignment shows large local variation even in the presence of strong image features.
We use this observation as an additional constraint in our robustness model.
In the case of large local motion variation -- computed as the length of local span of the displacement vectors magnitude -- we mark such region as likely having incorrect motion estimates:
\begin{equation}
\begin{split}
&\motionstr_x = \max_{j\in \neighbor{3}} \mv_x(j) - \min_{j\in \neighbor{3}} \mv_x(j),\\
&\motionstr_y = \max_{j\in \neighbor{3}} \mv_y(j) - \min_{j\in \neighbor{3}} \mv_y(j),\\
&\motionstr = \sqrt{\motionstr_x^2 + \motionstr_y^2},
\end{split}
\end{equation}
where $\mv_x$ and $\mv_y$ are horizontal and vertical displacements of the tile, $\motionstr_x$ and $\motionstr_y$ are local motion extents in horizontal and vertical direction in a $3 \times 3$ neighborhood $\neighbor{3}$, and $\motionstr$ is the final local motion strength estimation.
If $\motionstr$ exceeds a threshold value $\motionstrthreshold$ (see the supplemental material for details on the empirical tuning of $\motionstrthreshold$), we consider such pixel to be either containing significant local displacement or be misaligned and we use this information to scale up the robustness strength $\rejectionstrength$ (\Cref{eq:robustness}):
\begin{equation}
\rejectionstrength=\left\{\begin{matrix}
\rejectionstrength_1 & if \motionstr > \motionstrthreshold \\
\rejectionstrength_2 & otherwise
\end{matrix}\right.
\end{equation}

As a final step in our robustness computations, we perform additional refinement through morphological operations - we take a minimum confidence value in a $5 \times 5$ window:
\begin{equation}
\robustnessfinal=\min_{j\in \neighbor{5}} \robustness(j).
\end{equation}
This way we improve the robustness estimation in the case of misalignment in regions with high signal variance (like an edge on top of another one).

\begin{figure*}[t!]
  \centering
  \includegraphics[width=\linewidth]{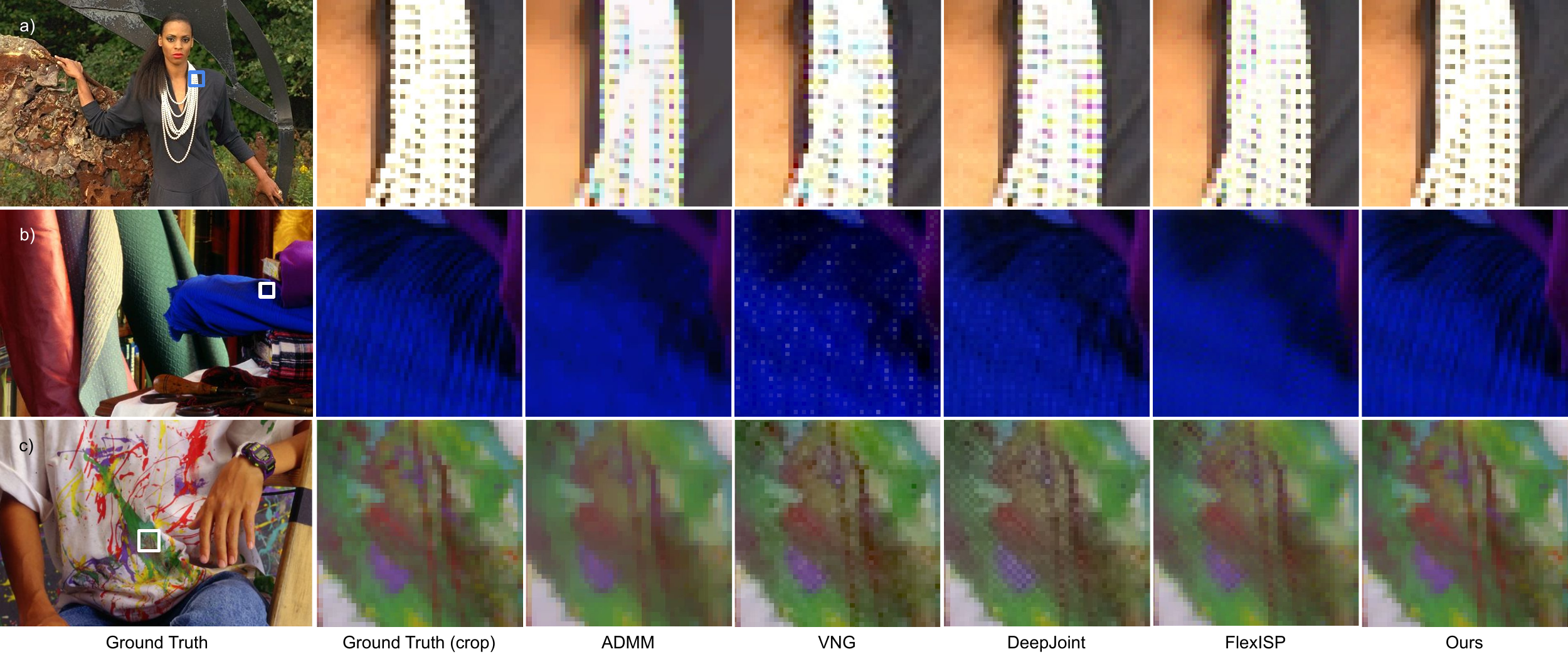}
  \caption{\textbf{Visual comparison of synthetic results.} Comparison of different single-image demosaicing techniques with that of our algorithm. Our algorithm uses information present between multiple frames to avoid typical demosaicing artifacts and is able to reconstruct most details in the case of highly saturated color areas.}
  \label{fig:result_synthetic_images}
\end{figure*}

\begin{figure*}[t!]
  \centering
  \includegraphics[width=\linewidth]{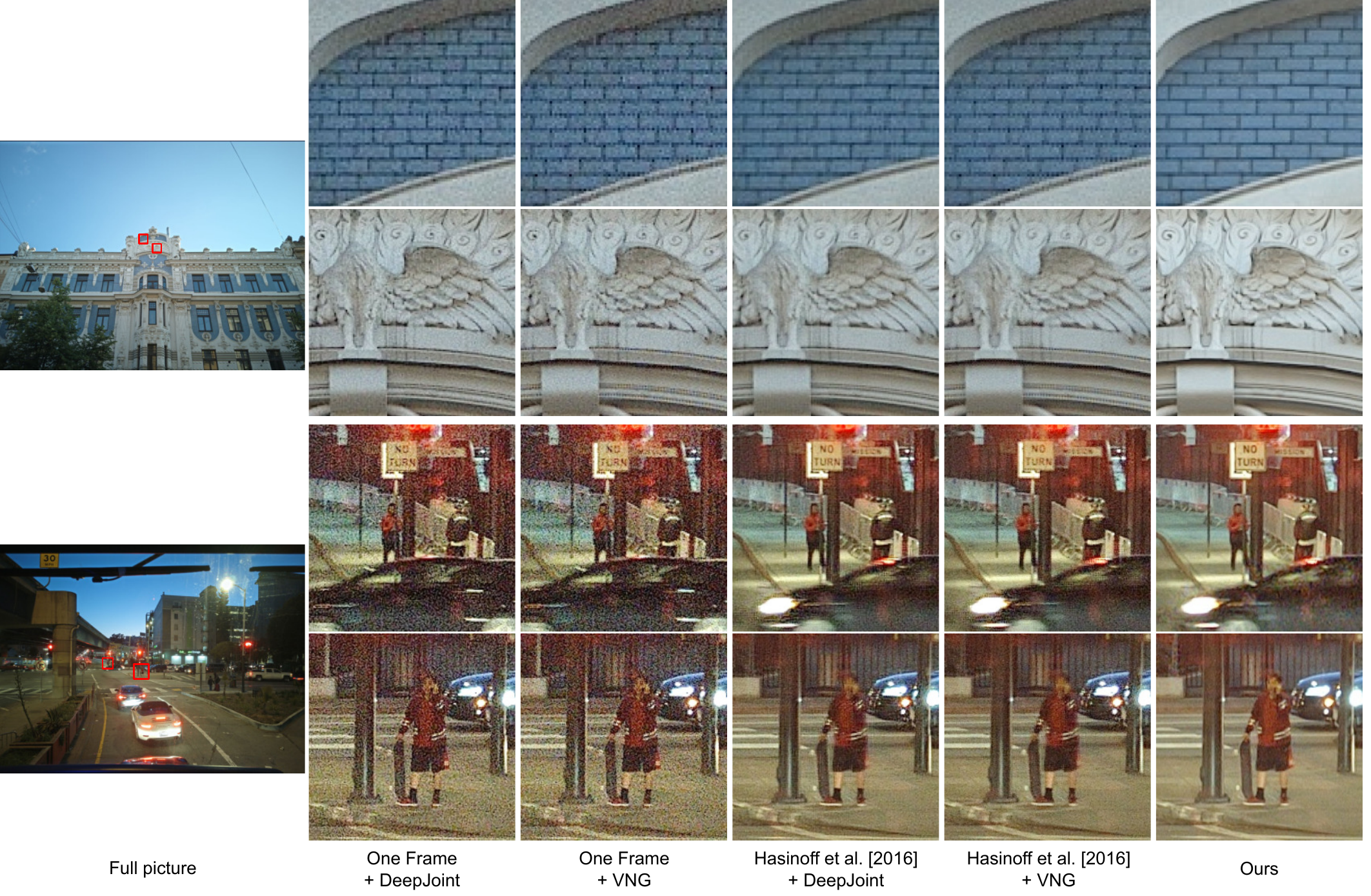}
  \vspace{-0.25in}
  \caption{\textbf{Comparison with demosaicing techniques:}
  Our method compared with dcraw's Variable Number of Gradients~\cite{Chang:1999:CFA} and \textit{DeepJoint}~\cite{Gharbi:2016:DJD}. Both demosaicing techniques are applied to either one frame from a burst or result of burst merging as described in Hasinoff \etal ~\shortcite{Hasinoff:2016:BPH}.
  Readers are encouraged to zoom in aggressively (300\% or more).}
  \label{fig:results_demosaicing}
\end{figure*}

\begin{figure*}[t!]
  \centering
  \includegraphics[width=\linewidth]{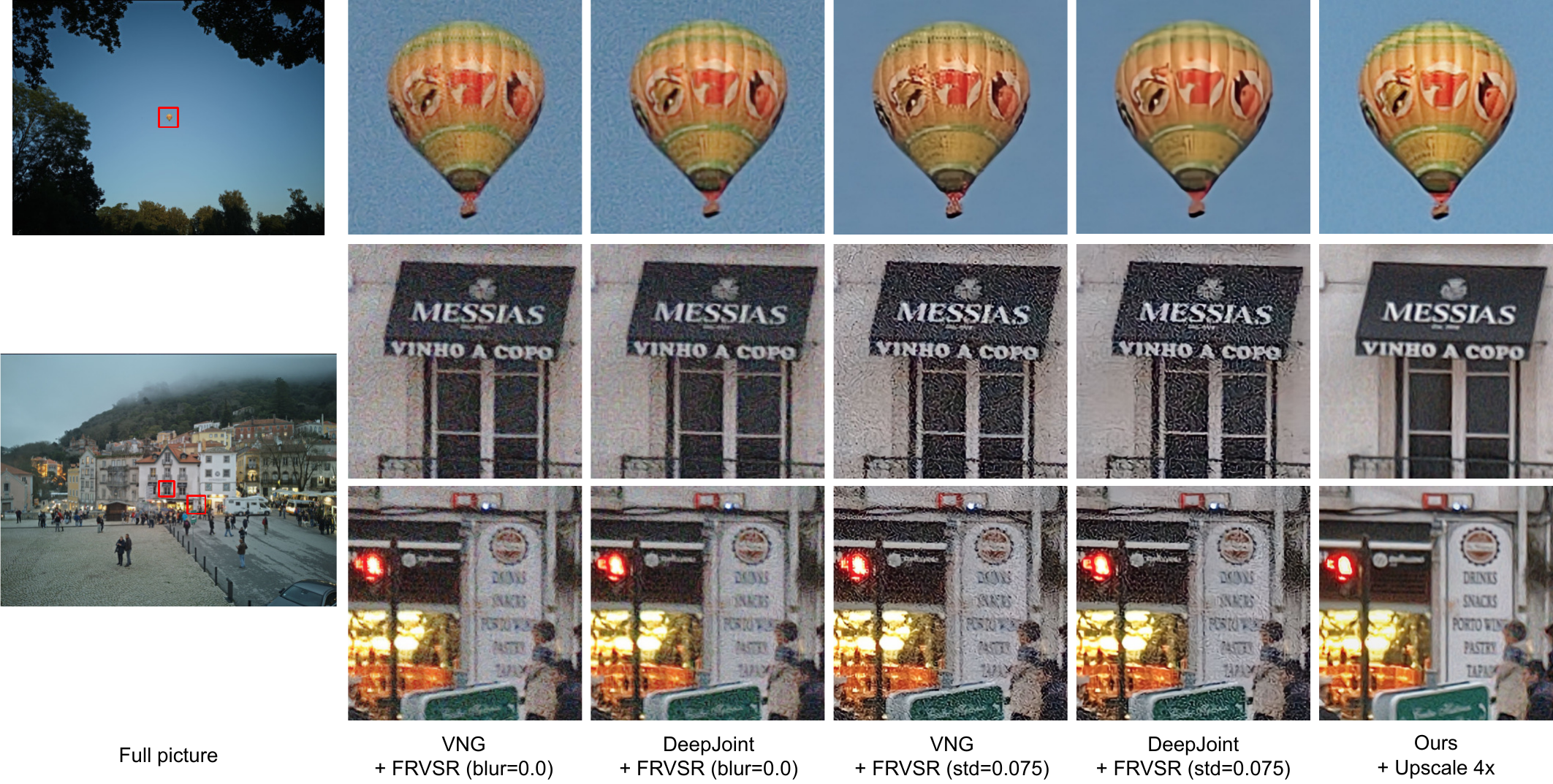}
  \caption{\textbf{Comparison with video super-resolution.}
  Our method compared with \textit{FRVSR}~\cite{Sajjadi:2018:FRVSR} applied to bursts of images demosaiced with VNG~\cite{Chang:1999:CFA} or \textit{DeepJoint}~\cite{Gharbi:2016:DJD}.
  Readers are encouraged to zoom aggressively (300\% or more).}
  \label{fig:video_sr_comparison_1}
\end{figure*}

\begin{figure*}[t!]
  \centering
  \includegraphics[width=\linewidth]{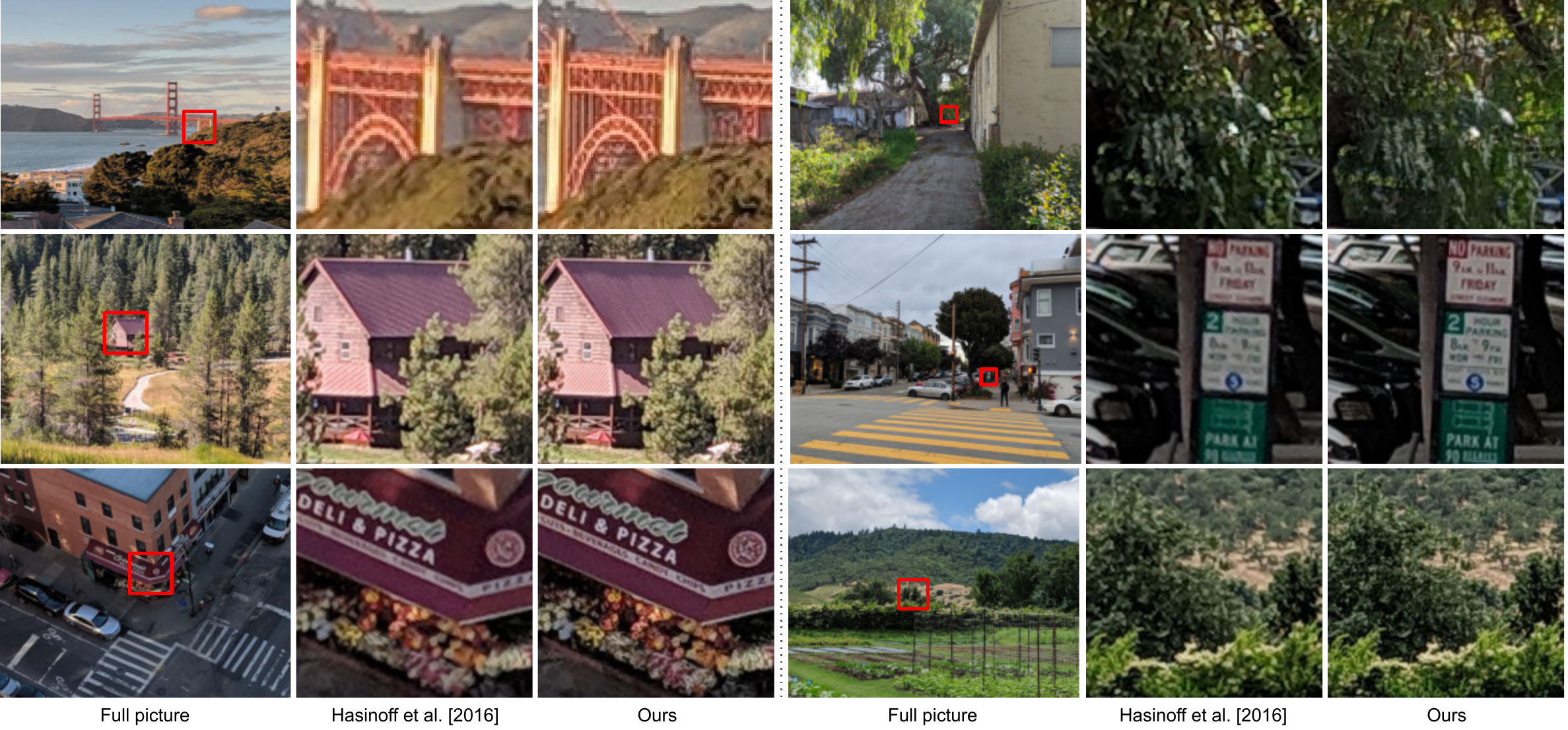}
  \caption{\textbf{End-to-end comparison as a replacement for the merge and demosaicing steps used in a camera pipeline.} Six bursts captured with a smartphone processed by a full camera pipeline described by Hasinoff \etal ~\shortcite{Hasinoff:2016:BPH}.
  Image crops on the left show results of merging using a temporal Wiener filter together with demosaicing, while image crops on the right show results of our algorithm.
  Our results show higher image resolution with more visible details and no aliasing effects.}
  \label{fig:results_end_to_end}
\end{figure*}

\begin{figure*}[t!]
  \centering
  \includegraphics[width=0.85\linewidth]{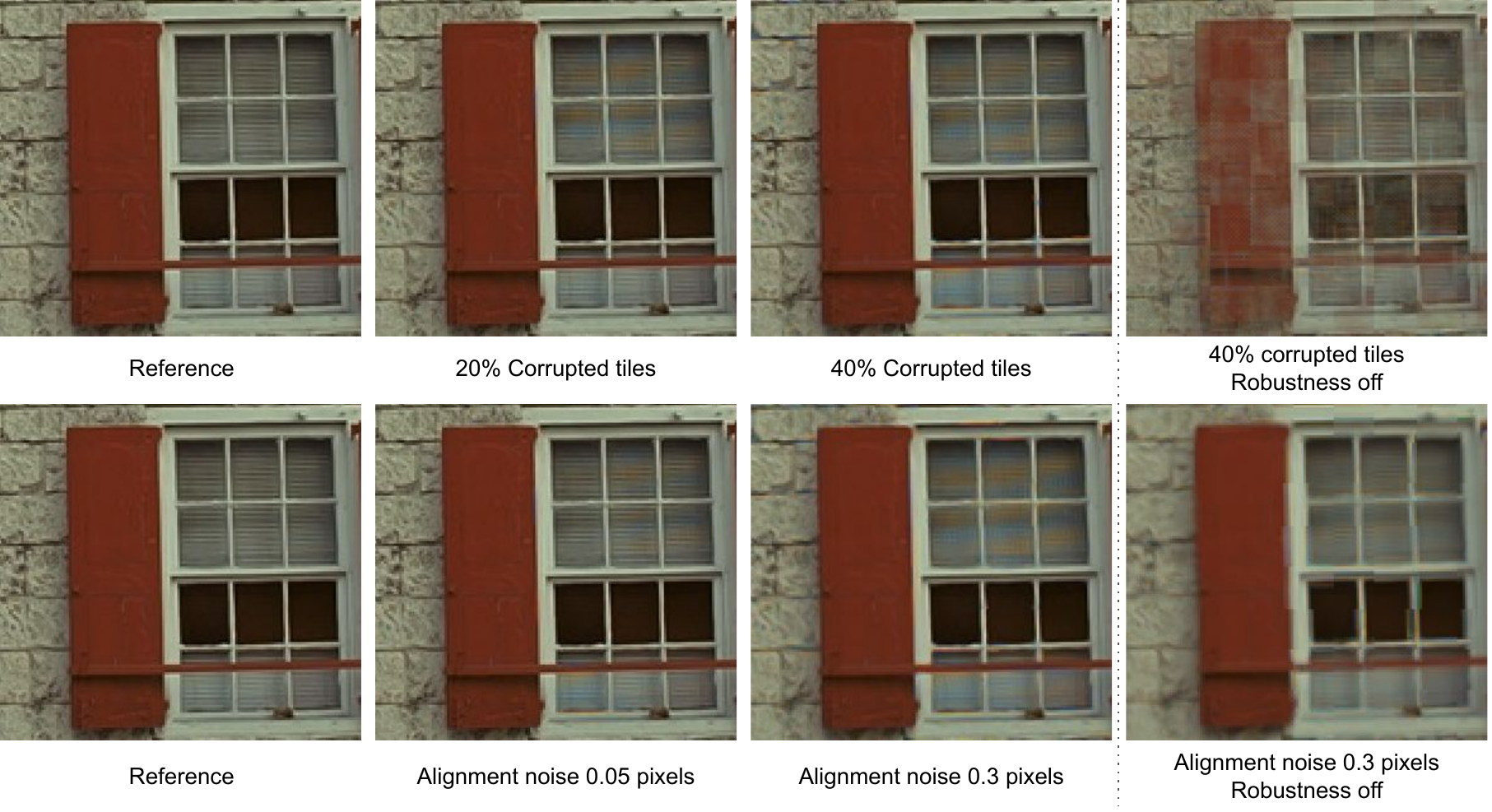}
  \caption{\textbf{Image quality degradation under misalignment.}
  \textbf{Top row:} Visual quality degradation caused by randomly corrupted and misaligned tiles.
  \textbf{Bottom row:} Visual quality degradation caused by noise added to the alignment vectors.
  From left to right we show the outputs corresponding to progressively more corrupted input data.
  The far-right example shows how the algorithm would behave without the motion robustness component.
  We observe that with increasing distortion rate, the algorithm tends to reject most frames, and results resemble a simple demosaicing algorithm with similar limitations and artifacts, but does not show fusion artifacts.
  }
  \label{fig:corruption_robustness_examples}
\end{figure*}

\begin{figure}[t!]
  \captionsetup{farskip=0pt}
  \centering
  \includegraphics[width=\linewidth]{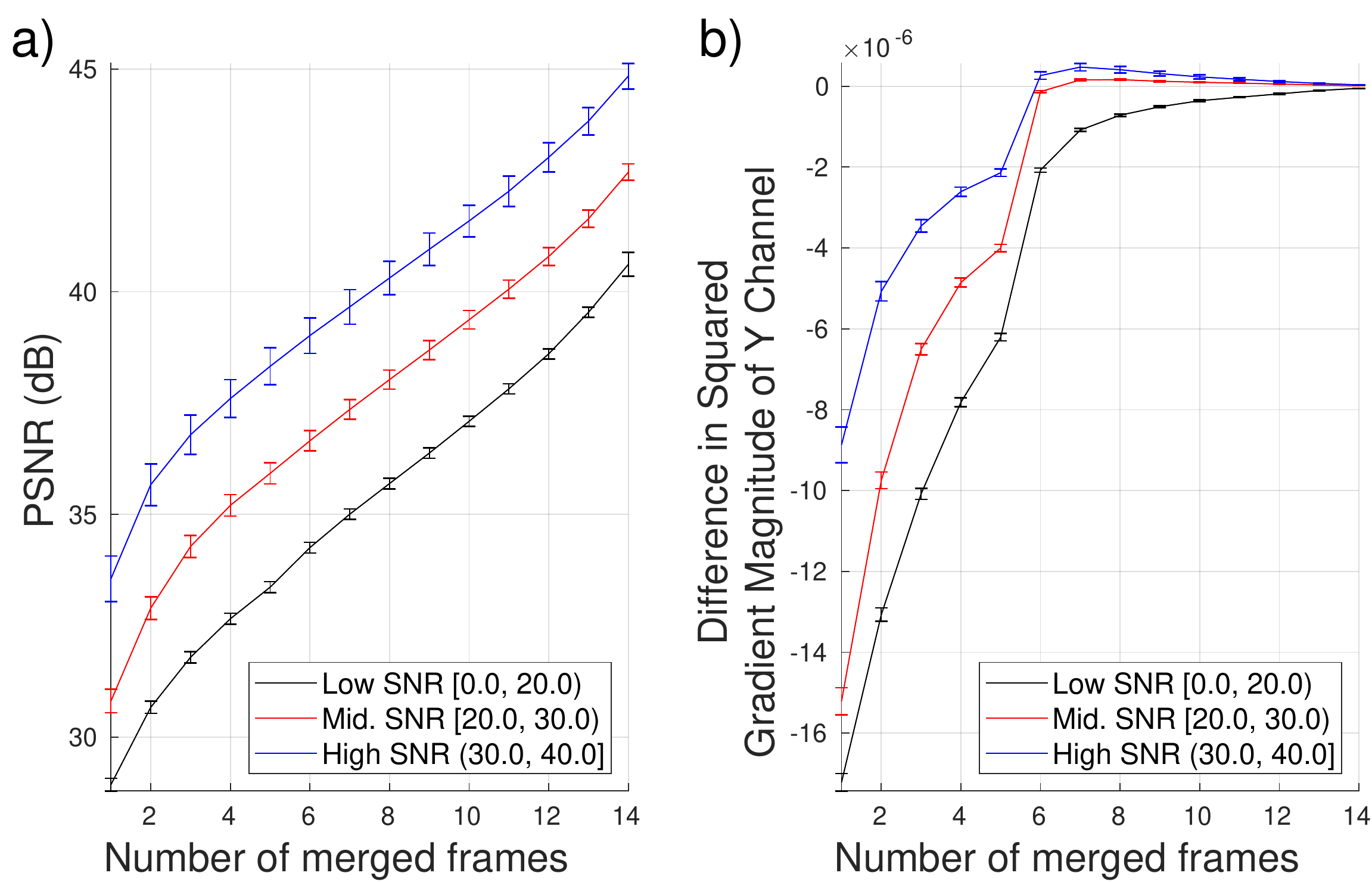}
  \caption{
  \textbf{(a)} PSNR results across different SNR ranges. The PSNR is measured between the result of merging $15$ frames to that of merging $n$ frames where $n=[1,..,14]$.
  \textbf{(b)} Plot presenting the sharpness (measured as average luminance gradient squared) difference between the result of merging $15$ frames to that of merging $n$ frames where $n=[1,..,14]$.
  We observe different behavior between low and high SNR of the base frame.  In good lighting conditions, sharpness reaches a peak around seven frames and then starts to degrade slowly.}
  \label{fig:quality_vs_n_frames}
\end{figure}

\begin{figure*}[t!]
  \centering
  \includegraphics[width=\linewidth]{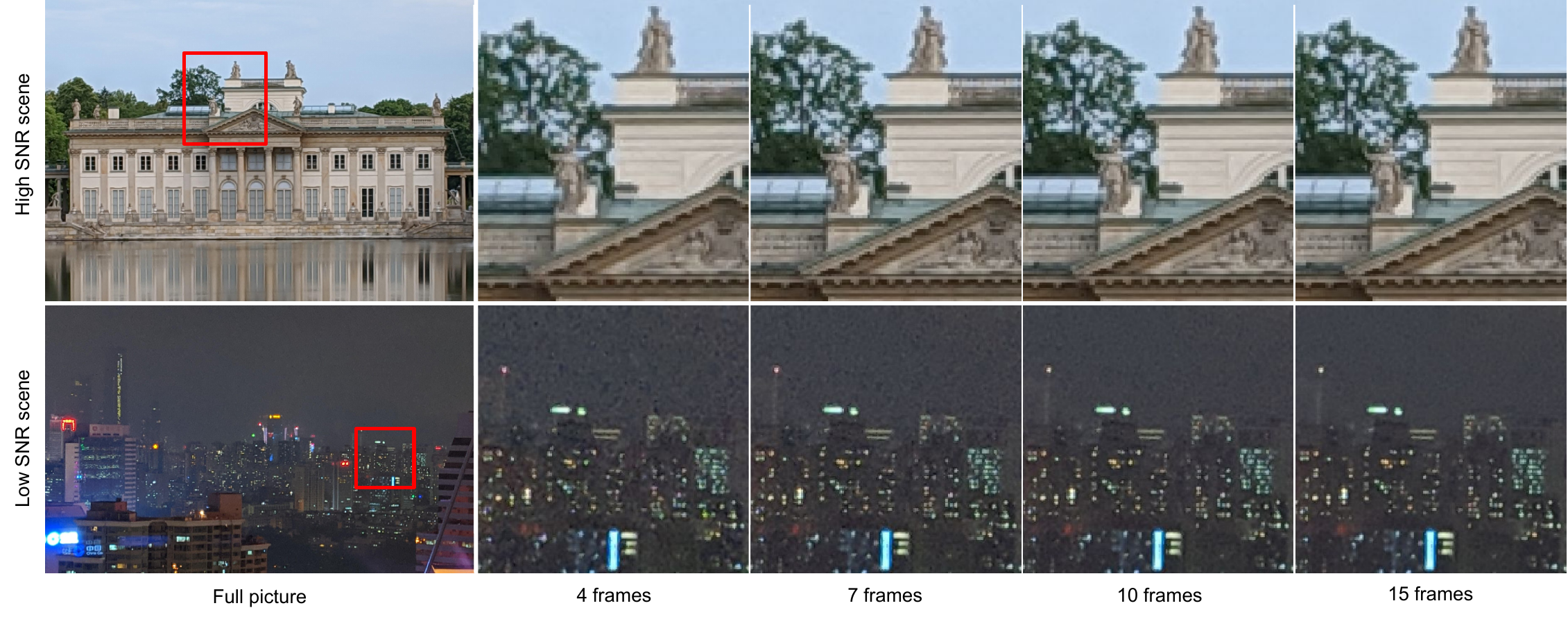}
  \caption{Visual differences caused by merging a different number of frames in the case of high \textbf{(top)} and low \textbf{(bottom)} SNR scenes.
  In the case the of high SNR scenes, we do not observe any image quality increase when using more than seven frames.
  On the other hand, in the case of low light scenes with worse SNR, we can observe a quality increase and better denoising.}
  \label{fig:different_frame_counts}
\end{figure*}

\section{Results}

To evaluate the quality of our algorithm, we provide the following analysis:
\begin{enumerate}
  \item Numerical and visual comparison against state of the art demosaicing algorithms using reference non-mosaic synthetic data.
  \item Analysis of the efficacy of our motion robustness after introduction of artificial corruption to the burst data.
  \item Visual comparison against state of the art demosaicing algorithms applied to real bursts captured by a mobile phone camera. We analyze demosaicing when applied to both a single frame and the results of the merge method described by Hasinoff \etal ~\shortcite{Hasinoff:2016:BPH}.
  \item End-to-end quality comparison inside a camera pipeline.
\end{enumerate}
We additionally investigate factors of relevance to our algorithm such as number of frames, target resolution, and computational efficiency.

\subsection{Synthetic Data Comparisons}\label{subsec:results_synthetic}

As we propose our algorithm as a replacement for the demosaicing step in classic camera pipelines, we compare against selected demosaicing algorithms:
\begin{itemize}
\item \textit{Variable Number of Gradients (VNG)}~\cite{Chang:1999:CFA} is used in the popular software open source processing tool \textit{dcraw}.
\item \textit{FlexISP} reconstructs images using a global optimization based on a single objective function ~\shortcite{Heide:2014:FAF}.
\item \textit{DeepJoint} Demosaicing and Denoising is a state-of-the-art neural network based approach by Gharbi \etal ~\shortcite{Gharbi:2016:DJD}.
\item \textit{ADMM} is an optimization based technique by Tan  \etal ~\shortcite{Tan:2017:JDD}.
\end{itemize}

To provide numerical full reference measurements (PSNR and SSIM), we need reference data. We create synthetic image bursts using two well-know datasets: Kodak (25 images) and McMaster (18 images) \cite{Zhang:2011:CDL}.
From each image, we created a synthetic burst, i.e., we generated a set of 15 random offsets (bivariate Gaussian distribution with a standard deviation of two pixels).
We resampled the image using nearest neighbor interpolation and created a Bayer mosaic (by discarding two of three color channels) to simulate the aliasing.
We measured the performance of our full algorithm against the direct single frame demosaicing techniques by comparing each algorithm's output to the original, non-resampled and non-Bayer image.

\Cref{table:kodak_psnr} presents the average PSNR and SSIM of all evaluated techniques on both datasets (box plots can be found in the supplemental material).
Our algorithm is able to use the information present across multiple frames and achieves the highest PSNR (over 3 dB better than the next best one, \textit{DeepJoint}, on both datasets) and SSIM numbers ($0.996$/$0.993$ vs $0.991$/$0.986$).
We additionally evaluate the results perceptually -- the demosaicing algorithms are able to correctly reproduce most of the original information with an exception of a few problematic areas.
In \Cref{fig:result_synthetic_images} we present some examples of demosaicing artifacts that our method avoids: color bleed (a), loss of details (b-c) and zipper artifacts (d).
In table \Cref{table:kodak_psnr} we also show the timings and computational performance of the used reference implementations.
As our algorithm was designed to run on a mobile device, it was highly optimized and uses a fast GPU processor, we achieve much better performance -- even when merging $15$ frames.

\begin{table}[h!]
\caption{\textbf{Quality analysis of our algorithm on synthetic data.}
Average PSNR and SSIM of selected demosaicing algorithms and our technique. Our algorithm has more information available across multiple frames and achieves the best quality results.}
\begin{tabular}{lccccc}
\hline
              & Kodak     & Kodak  & McM  & McM & MPix/s\\
              &  PSNR &  SSIM & PSNR & SSIM & (higher is better)\\ \hline
ADMM          & 31.79 & 0.935  & 32.66  & 0.957 & 0.0005 (CPU) \\
VNG           & 34.71 & 0.978  & 32.74  & 0.961 & 3.22 (CPU) \\
FlexISP       & 35.08 & 0.967  & 35.15  & 0.975 & 3.07 (GPU) \\
DeepJoint     & 39.67 & 0.991  & 37.58  & 0.986 & 0.33 (GPU) \\
\textbf{Ours} & \textbf{42.86} & \textbf{0.996} & \textbf{41.26} & \textbf{0.993} & 1756.9 (GPU) \\ \hline
\end{tabular}
\label{table:kodak_psnr}
\end{table}

\subsection{Motion robustness efficacy}\label{subsec:robustness_efficacy}
To evaluate the efficacy of motion robustness, we use the synthetic bursts generated in (\Cref{subsec:results_synthetic}) and additionally introduce two types of distortions.
The first distortion is to replace random alignment vectors with incorrect values belonging to a different area of the image.
This is similar to the behavior of real registration algorithms in case of significant movement and occlusion.
We corrupt this way with an increasing percentage of local alignment tiles $p=[10\%,..,50\%]$.

The second type of distortion is to introduce random noise to the alignment vectors, thereby shifting each image tile by a small, random amount.
This type of distortion is often caused by the noise or aliasing in the real images, or alignment ambiguity due to the aperture problem.
We add such noise independently to each tile and use normally distributed noise with standard deviation $\sigma=[0.05,..,0.25]$ displacement pixels.

Examples of both evaluations can be seen in (\Cref{fig:corruption_robustness_examples}).
Under very strong distortion, the algorithm fuses far fewer frames and behaves similarly to single-frame demosaicing algorithms.
While it shows similar artifacts to the other demosaicing techniques (color fringing, loss of detail), no multi-frame fusion artifact is present.
In the supplemental material, we include the PSNR analysis of the error with increasing amount of corruption, and more examples of how our motion robustness works on many different real bursts containing complex local motion or scene changes.

\subsection{Comparison on Real Captured Bursts}

We perform comparisons on real raw bursts captured with a Google Pixel~3 phone camera.
We compare against both single-frame demosaicing and the spatio-temporal Wiener-filter described by Hasinoff \etal ~\shortcite{Hasinoff:2016:BPH}, which also performs a burst merge.
As the output of all techniques is in linear space and is blurred by the lens, we sharpen it with an unsharp mask filter (with a standard deviation of three pixels) and apply global tonemapping -- an S-shaped curve with gamma correction.
We present some examples of this comparison in \Cref{fig:results_demosaicing}.
It shows that our algorithm produces the most detailed images with the least amount of noise.
Our results do not display artifacts like zippers in case of \textit{VNG} or structured pattern noise in case of \textit{DeepJoint}.
We show more examples in the supplemental material.

\begin{table*}
\caption{\textbf{User study:}
We used \emph{Amazon Mechanical Turk} and asked $115$ people - `Which camera quality do you prefer?'.
Shown, is a confusion matrix with the participants' nominated preference for each algorithm with respect to the other algorithms examined.
Each entry is the fraction of cases where the \emph{row} method was preferred relative to the \emph{column} method.
Our algorithm (highlighted in bold) was found to be preferred in more that $79\%$ of cases than that of the next best competing algorithm (Hasinoff et al. + DeepJoint).}
\begin{tabular}{lccccc}
    \hline
    ~                                    & One Frame          & One Frame         & Hasinoff et al.        & Hasinoff et al.        &  ~     \\
    ~                                    & DeepJoint          & VNG               & + DeepJoint            & VNG                    & Ours   \\ \hline
    One Frame + DeepJoint                & -                  & 0.495            & 0.198                 & 0.198                 & 0.059  \\
    One Frame + VNG                      & 0.505             & -                 & 0.099                  & 0.168                 & 0.059  \\
    Hasinoff et al. + DeepJoint          & 0.802             & 0.901            & -                      & 0.624                 & 0.208 \\
    Hasinoff et al. + VNG                & 0.802             & 0.832            & 0.376                 & -                      & 0.099  \\
    Ours                                 & \textbf{0.941}    & \textbf{0.941}   & \textbf{0.792}        & \textbf{0.901}        & -      \\ \hline
\end{tabular}
\vspace{-0.1in}
\label{table:user_study}
\end{table*}

We also conducted a user study to evaluate the quality of our method. For the study, we randomly generated four $250\times250$ pixel crops from each of the images presented in \Cref{fig:results_demosaicing}. To avoid crops with no image content, we discarded crops where the standard deviation of pixel intensity was measured to be less than $0.1$. In the study, we examined all paired examples between our method and the other compared methods. Using Amazon Mechanical Turk we titled each pair of crops, \textit{Camera A} and \textit{Camera B}, and asked 115 people to choose their prefered camera quality by asking the question - \textit{`Which camera quality do you prefer?'}. The crops were displayed at $50\%$ screen width and partipicants were limited to three minutes to review each example pair (the average time spent was $45$ seconds). \Cref{table:user_study} shows the participants' preferences for each of the compared methods. The results show that our method (in bold) is preferred in more than $79\%$ of cases than that of the next best method (Hasinoff et al. + DeepJoint).

As our algorithm is designed to be used inside a camera pipeline, we additionally evaluate the visual quality when our algorithm is used as a replacement of Wiener merge and demosaicing inside the~\cite{Hasinoff:2016:BPH} pipeline.
Some examples and comparisons can be seen in \Cref{fig:results_end_to_end}.

Finally, we compared our approach with \textit{FRVSR}~\cite{Sajjadi:2018:FRVSR}, state-of-the-art deep learning based video super-resolution.
We present some examples of this comparison in \Cref{fig:video_sr_comparison_1} (additional examples can be found in the supplemental material).
Note that our approach does not directly compete with this method since \textit{FRVSR}:
a. uses RGB images as the input (to be able to create the comparison we apply \textit{VNG} and \textit{DeepJoint} to the input Bayer images);
b. produces upscaled images (4x the input resolution) making a direct PSNR comparison difficult;
c. requires separate training for different light levels and amounts of noise present in the images.
In consultation with the authors of \textit{FRVSR}, we used two different versions of their network to enable the fairest possible comparison.
These different versions were not trained by us, but had been trained earlier and reported in their paper.
The versions of their network reported in our paper were the best performing for the given lighting and noise conditions: one trained on noisy input images with noise standard deviation of 0.075, the other on training data without any blur applied.
We show our method (after upscaling 4x~\cite{Romano:2017:RAISR}) compared with those two versions of \textit{FRVSR}.
In all cases, our method outperforms the combination of \textit{VNG} and \textit{FRVSR}.
In good light, our method shows amounts of detail comparable to the combination of \textit{DeepJoint} and \textit{FRVSR} at a fraction of their computational costs (\Cref{subsec:performance}), while in low light, our method shows more detailed images, no visual artifacts, and less noise.

\subsection{Quality Dependence on the Number of Merged Frames}\label{subsec:different_frame_count}
We use $586$ bursts captured with a mobile camera across different lighting to analyze the effect of merging different numbers of frames.
Theoretically, with the increase in the number of frames we have more information and observe better SNR.
However, the registration is imperfect, and the scene can change if the overall exposure time is too long.
Therefore, the inclusion of too many frames in the burst can diminish the quality of the results (\Cref{fig:quality_vs_n_frames} (b)).

The algorithm's current use of $15$ merged frames was chosen since it was found to produce high quality merged results from low-to-high SNR, and it was within the processing capabilities of a smartphone application.
We show in \Cref{fig:quality_vs_n_frames} (a) behavior of the PSNR as a function of $n$ merged frames for $n=[1,...,15]$ and across three different SNR ranges.
We observe approximately linear PSNR increase due to frame noise variance reduction with the increasing number of frames.
In case of good lighting conditions the imperceptible error of PSNR 40 dB is achieved around eight frames, while in low light the difference can be observed up to $15$ frames.
This behavior is consistent with the perceptual evaluation presented in \Cref{fig:different_frame_counts}.

\begin{figure*}[t!]
  \centering
  \includegraphics[width=0.95\linewidth]{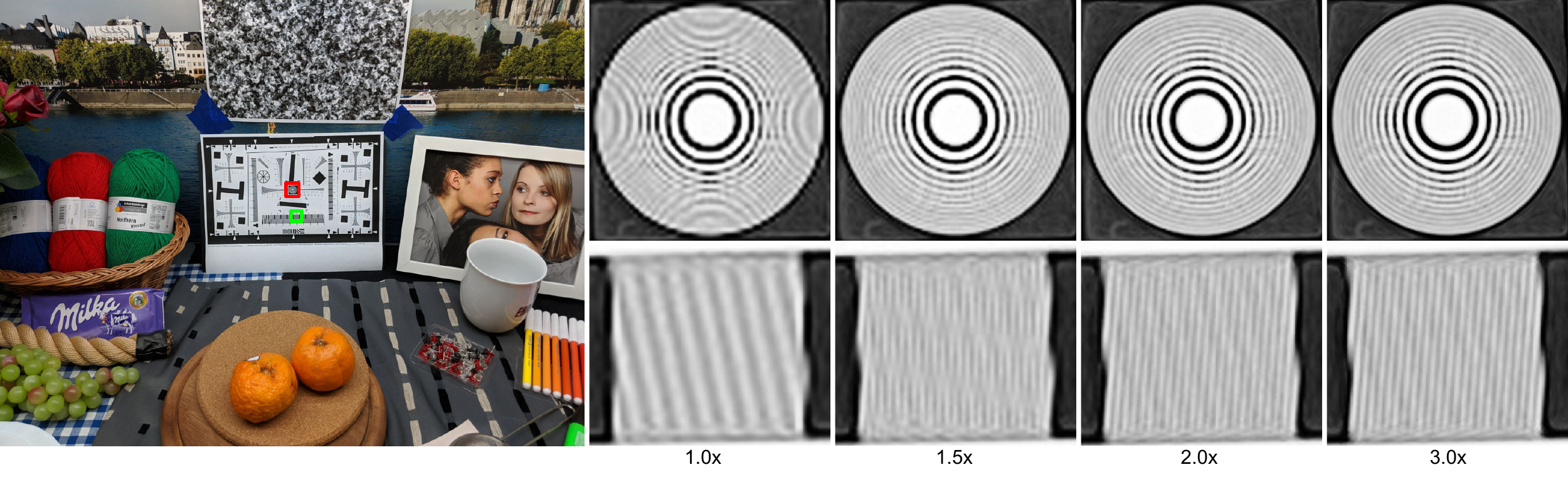}
  \vspace{-0.10in}
  \caption{\textbf{Different target grid resolutions.} Two different crops from a photo of a test chart, from left to right $1 \times$, $1.5 \times$, $2 \times$, and $3 \times$.
  Results were upscaled using a 3-lobe Lanczos filter to the same size. The combination of our algorithm and the phone's optical system with sampling ratio of $1.5$ leads to significantly improved results at $1.5 \times$ zoom, small improvement up to $2 \times$ zoom (readers are encouraged to zoom in) and no additional resolution gains returns thereafter.}
  \label{fig:different_grid_resolutions}
\end{figure*}

\begin{figure}[t!]
  \centering
  \includegraphics[width=\linewidth]{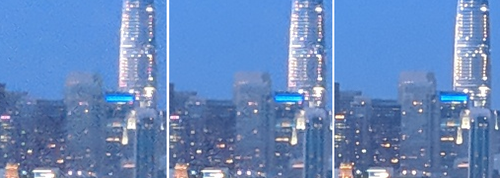}
  \vspace{-0.10in}
  \caption{\textbf{Comparison of behavior in low light.}
  \textbf{Left:} Single frame demosaiced and denoised with a spatial denoiser.
  \textbf{Middle:} Wiener filter spatio-temporal denoising of a burst ~\cite{Hasinoff:2016:BPH} followed by demosaicing.
  \textbf{Right:} Spatio-temporal denoising of our merge algorithm while preserving sharp local image details.}
  \label{fig:low_light}
\end{figure}

\begin{figure}[t!]
  \centering
  \includegraphics[width=\linewidth, trim=0cm 0cm 0cm 6cm, clip]{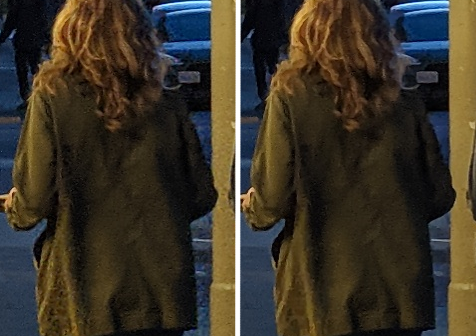}
  \vspace{-0.15in}
  \caption{\textbf{Occlusion and local motion in low light:}
  \textbf{Left:} Our motion robustness logic causes some regions to merge only single frame due to occlusions or misalignments.
  In low light it causes more noise in those regions.
  \textbf{Right:} Additional spatial denoising with strength inversely proportional to the merged frame count fixes those problems.}
  \label{fig:noisy_regions}
\end{figure}

\begin{table}[t!]
\caption{\textbf{Computational performance analysis of our algorithm.} We analyze timing and memory usage for two different hardware platforms, a mobile and a desktop GPU.
Timing cost comprises a fixed cost part at beginning and the end of the pipeline, while cost per frame grows linearly with the number of merged frames.
The cost scales linearly with the number of pixels. Runtime computational and memory cost makes our algorithm practical for use on a mobile device.}
\begin{tabular}{lccc}
\hline
GPU                & Fixed cost     & Cost per frame    & Memory cost  \\ \hline
\emph{Adreno 630}  & 15.4 ms        &  7.8 ms / MPix    & 22 MB / MPix \\
\emph{GTX 980}     &  0.83 ms       &  0.4 ms / MPix    & 22 MB / MPix \\ \hline
\end{tabular}
\label{table:performance}
\end{table}

\begin{figure}[ht]
  \centering
  \includegraphics[width=\linewidth]{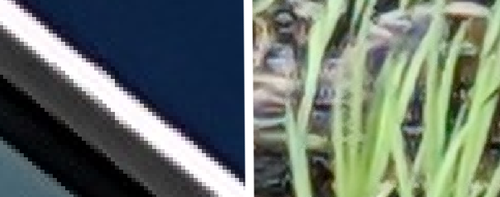}
  \caption{\textbf{Fusion artifacts:} \textbf{Left:} Aperture problem can create
  shifted object edges. \textbf{Right:} High frequency regions with subpixel motion
  can contribute to distinctive high frequency artifacts.}
  \label{fig:fusion_artifacts}
\end{figure}

\subsection{Computational Performance}\label{subsec:performance}
Our algorithm is implemented using OpenGL / OpenGL ES pixel shaders.
We analyze the computational performance of our method on both a Linux workstation with an \emph{nVidia GTX 980} GPU and on a mobile phone with a \textit{Qualcomm Adreno 630} GPU (included in many high-end 2018 mobile phones, including Google Pixel~3).
Performance and memory measurements to create a merged image can be found in \Cref{table:performance}.
They are measured per output image megapixel and scale linearly with the pixel count.
Because our algorithm merges the input images in an online fashion, the memory consumption is not dependent on the frame count.
The fixed initialization and finalizing cost is also not dependent on the frame count.
Those numbers indicate that our algorithm is multiple orders of magnitude faster than the neural network~\cite{Gharbi:2016:DJD} or optimization~\cite{Heide:2014:FAF} based techniques.

Similarly, our method is approximately two orders of magnitude faster as compared to \textit{FRVSR}~\cite{Sajjadi:2018:FRVSR} reported time of 191ms to process a single Full HD image on an nVidia P100 (10.5 MPix/s), even without taking into account the costs of demosaicing every burst frame.
Furthermore, the computational performance of our algorithm is comparable to that reported by Hasinoff \etal ~\shortcite{Hasinoff:2016:BPH} -- 1200 ms for just their merge technique in low light conditions, excluding demosaicing.

\section{Discussion and Limitations}\label{sec:discussion}
In this section, we discuss some of the common limitations of multi-frame SR approaches due to hardware, noise, and motion requirements.
We then show how our algorithm performs in some of the corner cases.

\subsection{Device Optics and Sampling}
By default our algorithm produces full RGB images at the resolution of the raw burst, but we can take it further.
The algorithm reconstructs a continuous representation of the image, which we can resample to the desired magnification and resolution enhancement factors.
The achievable super-resolution factor is limited by physical factors imposed by the camera design.
The most important factor is the \emph{sampling ratio}
\footnote{Ratio of the diffraction spot size of the lens to the number of pixels in the spot. A sampling ratio of two and above avoids aliasing.} at the focal plane, or sensor array.
In practical terms, this means the lens must be sharp enough to produce a relatively small lens spot size compared to the pixel size.
The sampling ratio for the main cameras on leading mobile phones such as the Apple iPhone~X and the Google Pixel~3 are in the range 1.5 to 1.8 (in the luminance channel) which are lower than the critical sampling ratio of two.
Since the sensor is color-filtered, the result Bayer raw images are aliased more -- the green channel is 50\% more aliased, where as the blue and red ones can be as much as twice more aliased.

We analyze super-resolution limits of our algorithm used on a smartphone using a raw burst captured with a Google Pixel~3 camera with a sampling ratio of approximately $1.5$.
We use different magnification factors ranging from $1 \times$ (just replacing the demosaicing step) up to $3 \times$ and use a handheld photo of a standard test chart.
\Cref{fig:different_grid_resolutions} presents a visual comparison between results achieved by running our algorithm on progressively larger target grid resolutions.
The combination of our algorithm and the phone's optical system leads to significantly improved results at $1.5 \times$ zoom, small improvement up to $2 \times$ zoom and no additional resolution gains returns thereafter.
Those results suggest that our algorithm is able to deliver resolution comparable to a dedicated tele lenses at modest magnification factors (under $2 \times$).

\subsection{Noise-dependent Tuning}
Beyond the optical/sensor design, there are also fundamental limits to super-resolution in the presence of noise.
This idea has been studied in a number of publications from both theoretical and experimental standpoints \cite{Helstrom:1969:DRI,Shahram:2006:SIT,Lu:2018:QOD}.
These works all note a power law relationship between achievable resolution and (linear) SNR.
Namely, the statistical likelihood of resolving two nearby point sources is proportional to \($SNR$^p\), where $0<p<1$ depends on the imaging system.
As SNR tends to zero, the ability to resolve additional details will also tend to zero rapidly.
Therefore, at low light levels, the ability to super-resolve is reduced, and most of the benefits of our algorithm will be manifested as spatio-temporal denoising that preserves the image details (\Cref{fig:low_light}).
Due to the difference between low and good light levels and the trade-offs it enforces (e.g., resolution enhancement vs. stronger noise reduction), we depend on the base frame signal-to-noise ratio for selecting the kernel tuning parameters.
The estimated SNR is the only input parameter to our tuning and is computed based on the average image brightness and the noise model \Cref{subsec:noisemodel}.
When presenting all of the visual results in this text, we have not manually adjusted any of the parameters per image and instead utilized this automatic SNR-dependent parameter selection technique.
The locally adaptive spatio-temporal denoising of our algorithm motivated its use as a part of the \textit{Night Sight} mode on Google's Pixel~3 smartphone.

\subsection{Lack of Movement}
As we highlighted in the main text, a key aspect of our approach is the reliance on random, natural tremor that is ubiquitously present in hand-held photography.
When the device is immobilized (for example when used on a tripod), we can introduce additional movement using active, moving camera components.
Namely, if the gyroscope on the device detects no motion, the sensor or the \emph{Optical Image Stabilization} system can be moved in a controlled pattern.
This approach has been successfully used in practice using sensor shifts (\textit{Sony A6000}, \textit{Pentax FF K1}, \textit{Olympus OM-D E-M1} or \textit{Panasonic Lumix DC-G9}) or the OIS movement~\cite{Wronski:2018:SBF}.

\subsection{Excessive Local Movement and Occlusion}
Our motion robustness calculation (\Cref{subsec:robustness}) excludes misaligned, moving or occluded regions from fusion to prevent visual artifacts.
However, in cases of severe local movement or occlusion, a region might get contributions only from the base frame, and our algorithm produces results resembling single frame demosaicing with significantly lower quality (\Cref{subsec:robustness_efficacy}).
In low light condition, these regions would also be much noiser than others, but additional localized spatial denoising can improve the quality, as demonstrated in \Cref{fig:noisy_regions}.

\subsection{Fusion Artifacts}\label{subsec:fusion_artifacts}
The proposed robustness logic (\Cref{subsec:robustness}) can still allow for specific minor fusion artifacts.
The alignment aperture problem can cause some regions to be wrongly aligned with similarly looking regions in a different part of the image.
If the difference is only subpixel, our algorithm could incorrectly merge those regions (\Cref{fig:fusion_artifacts} left).
This limitation could be improved by using a better alignment algorithm or a dedicated detector (we present one in the supplemental material).

Additionally, burst images may contain small, high frequency scene changes -- for example caused by ripples on the water or small magnitude leaf movement (\Cref{fig:fusion_artifacts} right).
When those regions get correctly aligned, the similarity between the frames makes our algorithm occasionally not able to distinguish those changes from real subpixel details and fuses them together.
Those problems have a characteristic visual structure and could be addressed by a specialized artifact detection and correction algorithm.

\section{Conclusions and Future Work}

In this paper we have presented a super-resolution algorithm that works on bursts of raw, color-filtered images.
We have demonstrated that given random, natural hand tremor, reliable image super-resolution is indeed possible and practical.
Our approach has low computational complexity that allows for processing at interactive rates on a mass-produced mobile device.
It does not require special equipment and can work with a variable number of input frames.

Our approach extends existing non-parametric kernel regression to merge Bayer raw images directly onto a full RGB frame, bypassing single-frame demosaicing altogether.
We have demonstrated (on both synthetic and real data) that the proposed method achieves better image quality than (a) state of the art demosaicing algorithms, and (b) state of the art burst processing pipelines that first merge raw frames and then demosaic (e.g. Hasinoff \etal ~\shortcite{Hasinoff:2016:BPH}).
By reconstructing the continuous signal, we are able to resample it onto a higher resolution discrete grid and reconstruct the image details of higher resolution than the input raw frames.
With locally adaptive kernel parametrization, and a robustness model, we can simultaneously enhance resolution and achieve local spatio-temporal denoising, making the approach suitable for capturing scenes in various lighting conditions, and containing complex motion.

An avenue of future research is extending our work to video super-resolution, producing a sequence of images directly from a sequence of Bayer images.
While our unmodified algorithm could produce such sequence by changing the anchor frame and re-running it multiple times, this would be inefficient and result in redundant computations.

For other future work, we note that computational photography, of which this paper is an example, has gradually changed the nature of photographic image processing pipelines.
In particular, algorithms are no longer limited to pixel-in / pixel-out arithmetic with only fixed local access patterns to neighboring pixels.
This change suggests that new approaches may be needed for hardware acceleration of image processing.

Finally, depending on handshake to place red, green, and blue samples below each pixel site suggest that perhaps the design of color filter arrays should be reconsidered; perhaps the classic RGGB Bayer mosaic is no longer optimal.
Perhaps the second G pixel can be replaced with another sensing modality.
More exotic CFAs have traditionally suffered from reconstruction artifacts, but our rather different approach to reconstruction might mitigate some of these artifacts.

\begin{acks}
We gratefully acknowledge current and former colleagues from collaborating teams across Google including: Haomiao Jiang, Jiawen Chen, Yael Pritch, James Chen, Sung-Fang Tsai, Daniel Vlasic, Pascal Getreuer, Dillon Sharlet, Ce Liu, Bill Freeman, Lun-Cheng Chu, Michael Milne, and Andrew Radin.
Integration of our algorithm with the Google Camera App as \textit{Super-Res Zoom} and \textit{Night Sight} mode was facilitated with generous help from the Android camera team.
We also thank the anonymous reviewers for valuable feedback that has improved our manuscript.
\end{acks}

\bibliographystyle{ACM-Reference-Format}
\bibliography{handheld_multiframe_superres_sig2019}

\ifthenelse{\equal{\mergewithsuppement}{1}}
{
\cleardoublepage
\newcommand{\beginsupplement}{%
        \renewcommand{\thetable}{\arabic{table}}%
        \renewcommand{\thefigure}{\arabic{figure}}%
        \renewcommand{\thesection}{S}%
        \renewcommand{\thesubsection}{S.\arabic{subsection}}%
}

\pagenumbering{roman}
\beginsupplement
\section{SUPPLEMENT}
\subsection{Adaptive Super-Resolution and Denoising}\label{subsec:denoising}
\begin{figure}[tbh]
  \centering
  \includegraphics[width=\linewidth]{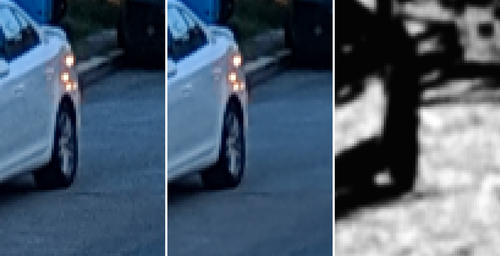}
  \caption{\textbf{Denoising:} Example effect of local kernel denoising,
  \textbf{Left}: Low light image without local kernel denoising, $\kerneldenoise = 1.0$.
  \textbf{Middle}: image with strong local kernel denoising $\kerneldenoise = 5.0$.
  \textbf{Right}: local denoising mask.
  Black pixels denote areas where we do not apply any spatial denoising and adjust kernel values for super-resolution, while white pixels denote areas where we do not observe enough image details to justify super-resolution and adjust the kernel values for denoising.
  By analyzing the local structure, our algorithm can cover a continuous balance between resolution enhancement and spatio-temporal denoising.}
  \label{fig:fig_local_kernel_denoising}
\end{figure}

In Section 5.1.2 we describe adapting the spatial support of the sampling kernel based on the local gradient structure tensor.
We use the magnitude of the structure tensor's dominant eigenvalue $\eigenval_1$ to drive the spatial support of the kernel and the trade-off between the super-resolution and denoising, where $\frac{\eigenval_1-\eigenval_2}{\eigenval_1+\eigenval_2}$ is used to drive the desired anisotropy of the kernels (Figure 7 in the main paper text).
We use the following heuristics to estimate the kernel shapes ($\kernel_1$ and $\kernel_2$ in Equation (4) in the main paper text):
\begin{equation*}
\begin{split}
&\anisotropy = 1 + \sqrt\frac{\eigenval_1-\eigenval_2}{\eigenval_1+\eigenval_2},\\
&\denoise = clamp(1 - \frac{\sqrt{\eigenval_1}}{\denoisetransition} + \denoisethreshold, 0, 1),\\
&\hat{\kernel_1} = \kernelbase \cdot (\kernelstretch \cdot \anisotropy),\\
&\hat{\kernel_2} = \frac{\kernelbase}{(\kernelshrink \cdot \anisotropy)},\\
&\kernel_1 = ((1 - \denoise) \cdot \hat{\kernel_1} + \denoise \cdot \kernelbase \cdot \kerneldenoise)^2,\\
&\kernel_2 = ((1 - \denoise) \cdot \hat{\kernel_2} + \denoise \cdot \kernelbase \cdot \kerneldenoise)^2.
\end{split}
\end{equation*}

We use the symbol $\anisotropy$ for the computed gradient anisotropy and $\denoise$ for the estimated denoising strength.
We use the following tuning parameters: $\denoisethreshold$ as the denoising threshold, $\denoisetransition$ as how fast we go from full denoising to no denoising depending on the gradient strength, $\kernelstretch$ as the amount of kernel stretching along the edges, $\kernelshrink$ as the amount of kernel shrinking perpendicular to the edges, $\kernelbase$ as the base kernel standard deviation, and $\kerneldenoise$ as the kernel standard deviation suitable for denoising.
The denoising strength will make the whole kernel shape bigger and more radial, effectively also overriding the anisotropic stretching in regions that are candidates for denoising.

The reasoning behind these heuristics is that small dominant eigenvalues (comparable to the amount of noise expected in the given raw image) signify relatively flat, noisy regions while large eigenvalues appear around features whose resolution we want to enhance (\Cref{fig:fig_local_kernel_denoising}).
\Cref{fig:fig_local_kernel_denoising} \textbf{left} and \textbf{middle} show the visual impact of $\kerneldenoise$ parameter, while the contrast of the mask presented on the \textbf{right} depends on $\denoisethreshold$ and $\denoisetransition$.

\subsection{Tuning Procedure and Parameters}\label{subsec:tuning}
In this section we describe the tuning parameters that we used for the results presented for our algorithm.
Parameters that affect the trade-off between the resolution-increase and spatio-temporal denoising (\Cref{subsec:denoising}) depend on the signal-to-noise ratio of the input frames.
In such case the parameters are piece-wise linear functions of SNR in the range $[6..30]$.
\begin{equation*}
\begin{split}
&\tilesize = [16,32,64]px, \\
&\kernelbase = [0.25,...,0.33]px, \\
&\kerneldenoise = [3.0,...,5.0], \\
&\denoisethreshold = [0.001,...,0.010], \\
&\denoisetransition = [0.006,...,0.020], \\
&\kernelstretch = 4, \\
&\kernelshrink = 2, \\
&\rejectionthreshold = 0.12,\\
&\rejectionstrength_1 = 12,\\
&\rejectionstrength_2 = 2,\\
&\motionstrthreshold = 0.8px.\\
\end{split}
\end{equation*}
The $\tilesize$, $\kernelbase$, and $\motionstrthreshold$ are in units of pixels, $\denoisethreshold$ and $\denoisetransition$ are in units of gradient magnitude of the image normalized to the range $[0,...,1]$.
The remaining parameters are either unitless multipliers ($\kerneldenoise$, $\kernelstretch$, $\kernelshrink$) or operate on color differences normalized by the standard deviation ($\rejectionthreshold$, $\rejectionstrength_1$, $\rejectionstrength_1$).

Since our algorithm is designed to produce visually pleasing images taken with a mobile camera, we tuned those parameters based on perceputal image quality assessment ensuring visual consistency for SNR values from 6 to over 30 where the SNR was measured from a single frame.
Next, we discuss the impact of some of those parameters on the final image.
\begin{figure}[tbh]
  \centering
  \includegraphics[width=1.0\linewidth]{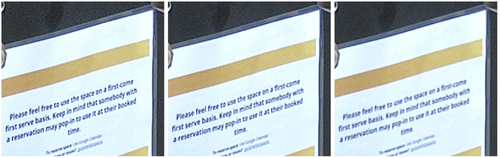}
  \caption{\textbf{Impact of $\kernelbase$ on the visual results.}
  \textbf{Left:} $\kernelbase$ of 0.1px produces very sharp results with significant amounts of noise and some artifacts.
  \textbf{Middle:} $\kernelbase$ of 0.25px produces results balanced between resolution enhancement and denoising.
  \textbf{Right:} $\kernelbase$ of 0.4px produces over-smoothed results.
  }
  \label{fig:base_kernel_size}
\end{figure}
The chosen kernel parameters balance the resolution enhancement with suppression of noise and artifacts in the image. \Cref{fig:base_kernel_size} shows the visual impact of adjusting the base kernel size $\kernelbase$.
Figure 7 presented earlier in the main paper shows how the $\kernelstretch$ and $\kernelshrink$ impact the result, smoothing the edges and getting rid of alignment artifacts that can result from the aperture problem.
The $\tilesize$ is increased from $16px$ to up to $64px$ in very low light situations to increase the robustness of alignment to significant amounts of noise.

\begin{figure}[tbh]
  \centering
  \includegraphics[width=0.9\linewidth]{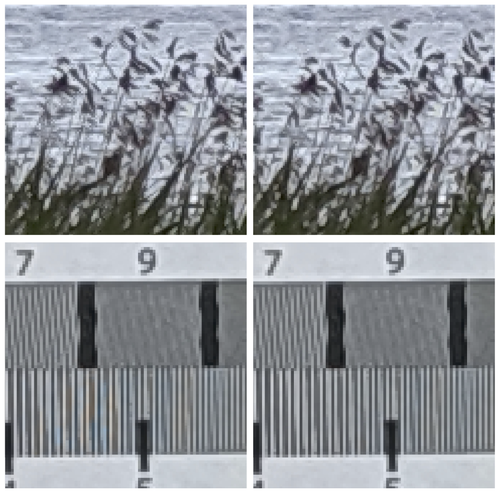}
  \caption{\textbf{Impact of $\rejectionstrength_2$ on the visual results.}
  \textbf{Top-left:} Too small $\rejectionstrength_2$ of 1 produces small high-frequency artifacts.
  \textbf{Bottom-left:} Too large $\rejectionstrength_2$ of 4 causes over-rejection in highly aliased regions and loss of super-resolution.
  \textbf{Bottom-right and top-right:} $\rejectionstrength_2$ of 2 correctly treats areas with local movement as well as heavily aliased regions.
  }
  \label{fig:rejection_tuning}
\end{figure}

Tuning of the $\rejectionstrength$ and $\motionstrthreshold$ is performed to balance the false-positive and the false-negative rate of our robustness logic.
A rejection rate that is too large leads to not merging some heavily aliased areas (like test chart images), while too small rejection rate leads to the manifestation of fusion artifacts.
The effect of having this parameter too small or too large can be observed in \Cref{fig:rejection_tuning}.
In practice, to balance those effects, we use the same fixed values for all processed images.

\begin{figure}[tbh]
  \centering
  \includegraphics[width=\linewidth]{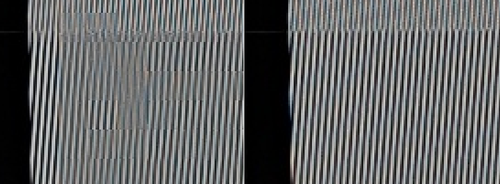}
  \caption{\textbf{High frequency artifacts caused by the aperture problem:} \textbf{Left:} a high resolution and high frequency test chart image without the rejection logic described in ~\Cref{subsec:aperture}.
  Notice the numerous blocky artifacts visible when zoomed-in. \textbf{Right:} the same image with the rejection logic detecting variance loss showing no fusion artifacts, but some aliasing and color fringing.}
  \label{fig:fig_unblocker_2}
\end{figure}

\subsection{High Frequency Artifacts Removal}\label{subsec:aperture}
Alignment algorithms (such as block matching or gradient based) fail to correctly align high frequency repetitive patterns (due to the aperture problem).
Our robustness logic makes use of both low-pass filtering and comparing local statistics.
Therefore, the algorithm as described is prone to producing blocky artifacts in regions containing \emph{only} very high frequency signals, often observed on human-made test charts (\Cref{fig:fig_unblocker_2}).
To prevent this effect, we detect those regions by analyzing the local variance loss caused by local lowpass filtering.
In particular, we compare the local variance before and after the lowpass filtering.
When we detect variance loss and a large local variation in the alignment vector field (the same as used in the motion prior in Section 5.2.3), we mark those regions as incorrectly aligned and fully reject them.
An example comparison with and without this logic is presented in \Cref{fig:fig_unblocker_2}.
This heuristic has a trade-off: in some cases, even properly aligned high frequency regions do not get merged.

\subsection{Synthetic Data Quality Analysis}
We show detailed box plots of our algorithm's performance compared to different demosaicing techniques in \Cref{fig:kodak_psnr}.
\begin{figure}[t!]
  \centering
  \includegraphics[width=\linewidth]{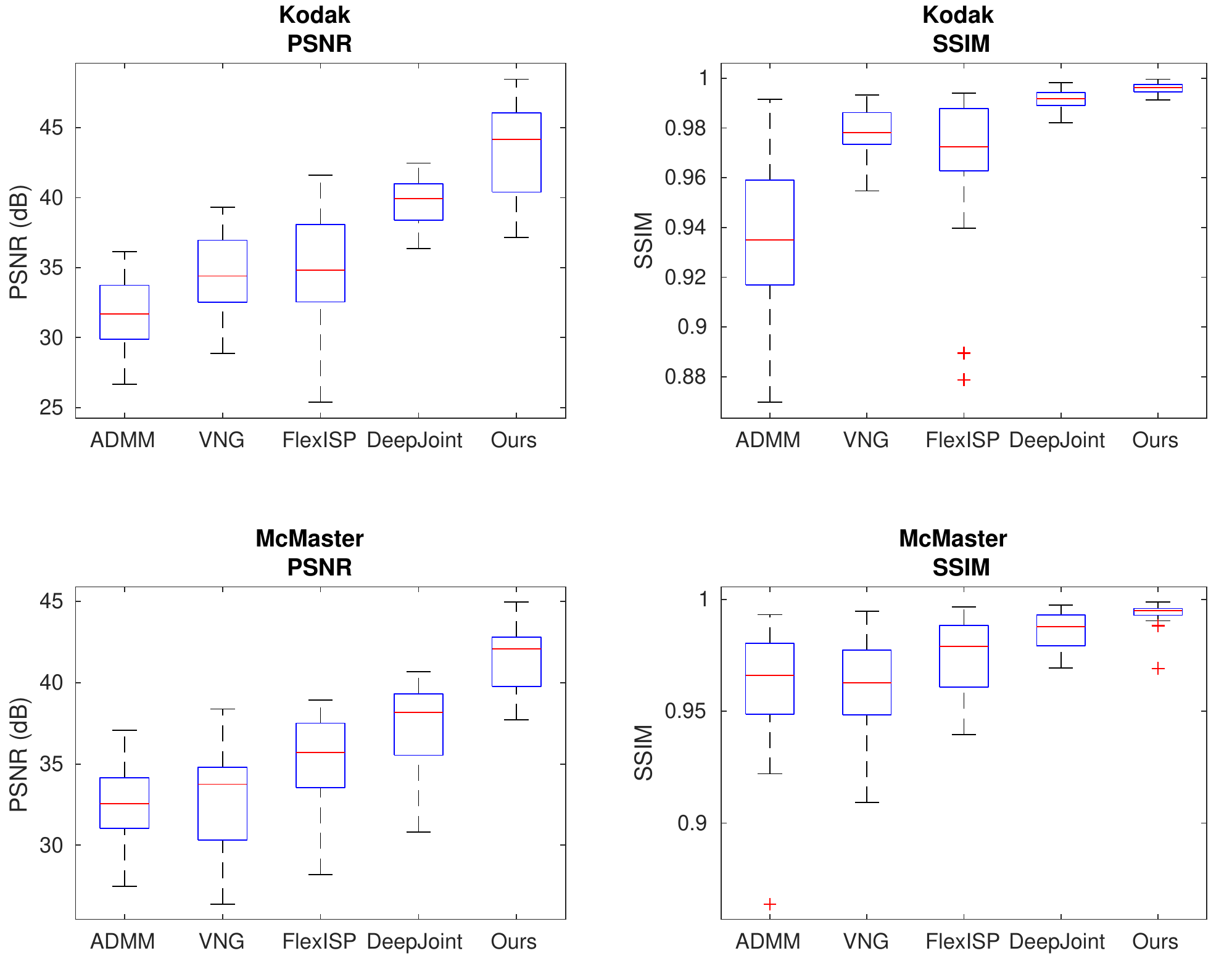}
  \caption{\textbf{PSNR and SSIM comparisons on Kodak and McMaster dataset.}
  Performance of our algorithm compared to alternate approaches using PSNR and SSIM on synthetic bursts created from the Kodak and McMaster datasets.
  Our solution can use information present across multiple frames and is significantly better than all other techniques on both synthetic datasets.}
  \label{fig:kodak_psnr}
\end{figure}

\subsection{Robustness Analysis}
A PSNR analysis of the robustness on synthetic alignment corruption tests is shown in \Cref{fig:corruption_robustness}.
The strongest quality degradation (50\% corrupted image tiles or wrong alignment with random offsets of $0.25$ pixels) leads to our algorithm merging only a single frame and PSNR values comparable to simple demosaicing techniques.
Additionally, we show examples of burst merging with and without the robustness model in real captured bursts in different difficult conditions in \Cref{fig:robustness_supplement}.

\begin{figure}[t!]
  \centering
  \includegraphics[width=1.0\linewidth]{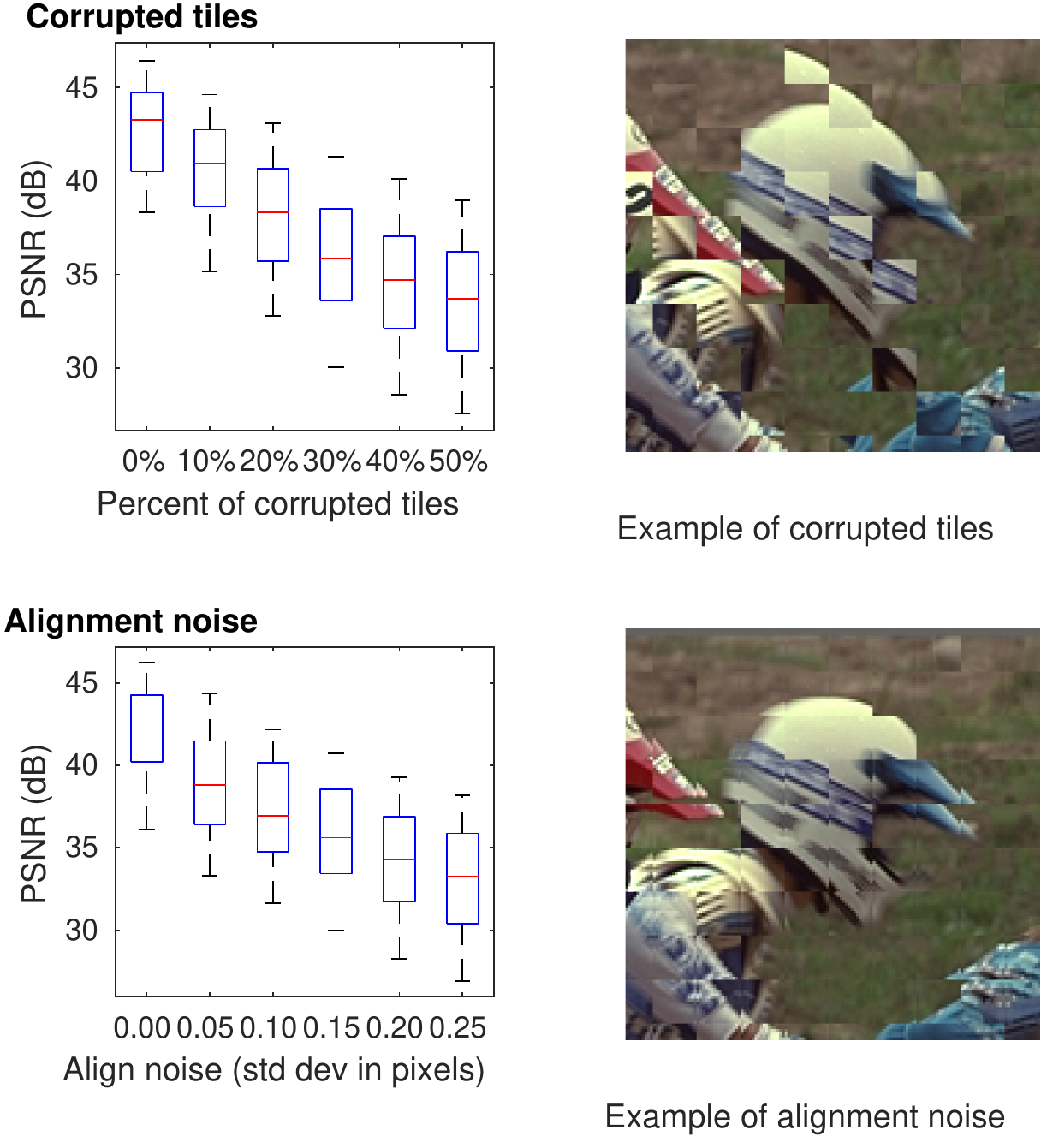}
  \caption{\textbf{PSNR of image quality caused by alignment corruption of synthetic bursts created from Kodak dataset.}
  \textbf{Top-Left:} PSNR of our algorithm output caused by randomly corrupted and misaligned tiles.
  \textbf{Bottom-Left:} Visual demonstration of this type of distortion at the highest evaluated distortion value.
  \textbf{Top-Right:} PSNR of our algorithm output caused by noise added to the alignment vectors.
  \textbf{Bottom-Right:} Visual demonstration of this type of distortion at the highest evaluated distortion value.
  With increasing distortion rate we observe gradual quality degradation, as our algorithm rejects most of the frames in the synthetic burst and degrades to a simple gradient-based demosaicing technique.
  }
  \label{fig:corruption_robustness}
\end{figure}

\begin{figure*}[tbh]
  \centering
  \includegraphics[width=\linewidth]{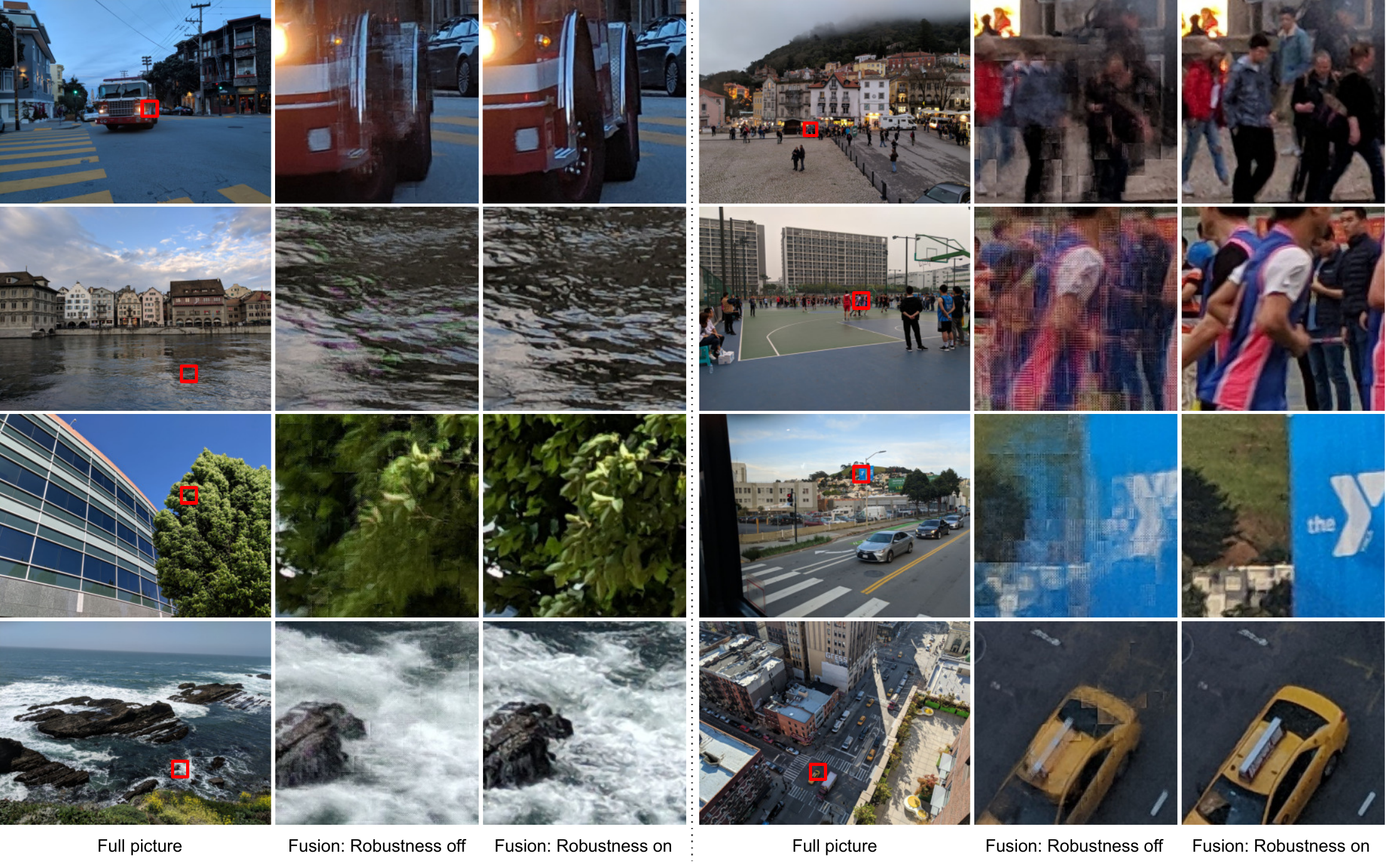}
  \caption{\textbf{Robustness examples:}
    \textbf{Left:} Full photo.
    \textbf{Middle:} Crop of the photo merged without our robustness model.
    \textbf{Right:} Same region of the photo merged with our robustness model.
    In real captured bursts, our algorithm is able to handle challenging scenarios including local scene motion, parallax or scene changes like water rippling.}
  \label{fig:robustness_supplement}
\end{figure*}

\subsection{Real Captured Bursts Additional Results}
We show some additional comparisons with competing techniques on bursts captured with a mobile camera in \Cref{fig:results_demosaicing_supplement}.

\begin{figure*}[tbh]
  \centering
  \includegraphics[width=\linewidth]{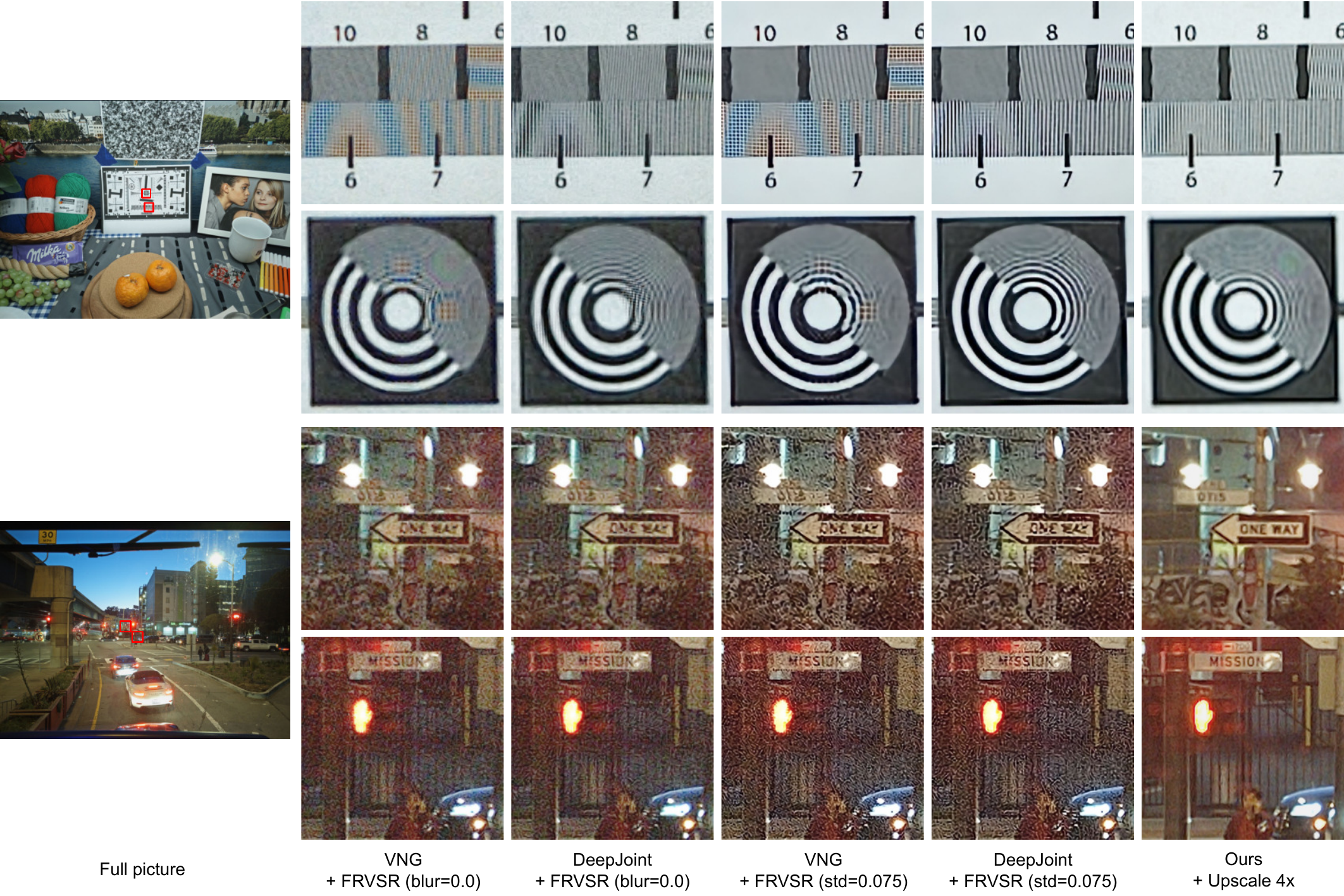}
  \caption{\textbf{Additional comparison with video super-resolution.}
  Our method compared with \textit{FRVSR}~\cite{Sajjadi:2018:FRVSR} applied to bursts of images demosaiced with VNG~\cite{Chang:1999:CFA} or \textit{DeepJoint}~\cite{Gharbi:2016:DJD}.
  Readers are encouraged to zoom aggressively (300\% or more).}
  \label{fig:video_sr_comparison_2}
\end{figure*}

\begin{figure*}[t!]
  \centering
  \includegraphics[width=0.9\linewidth]{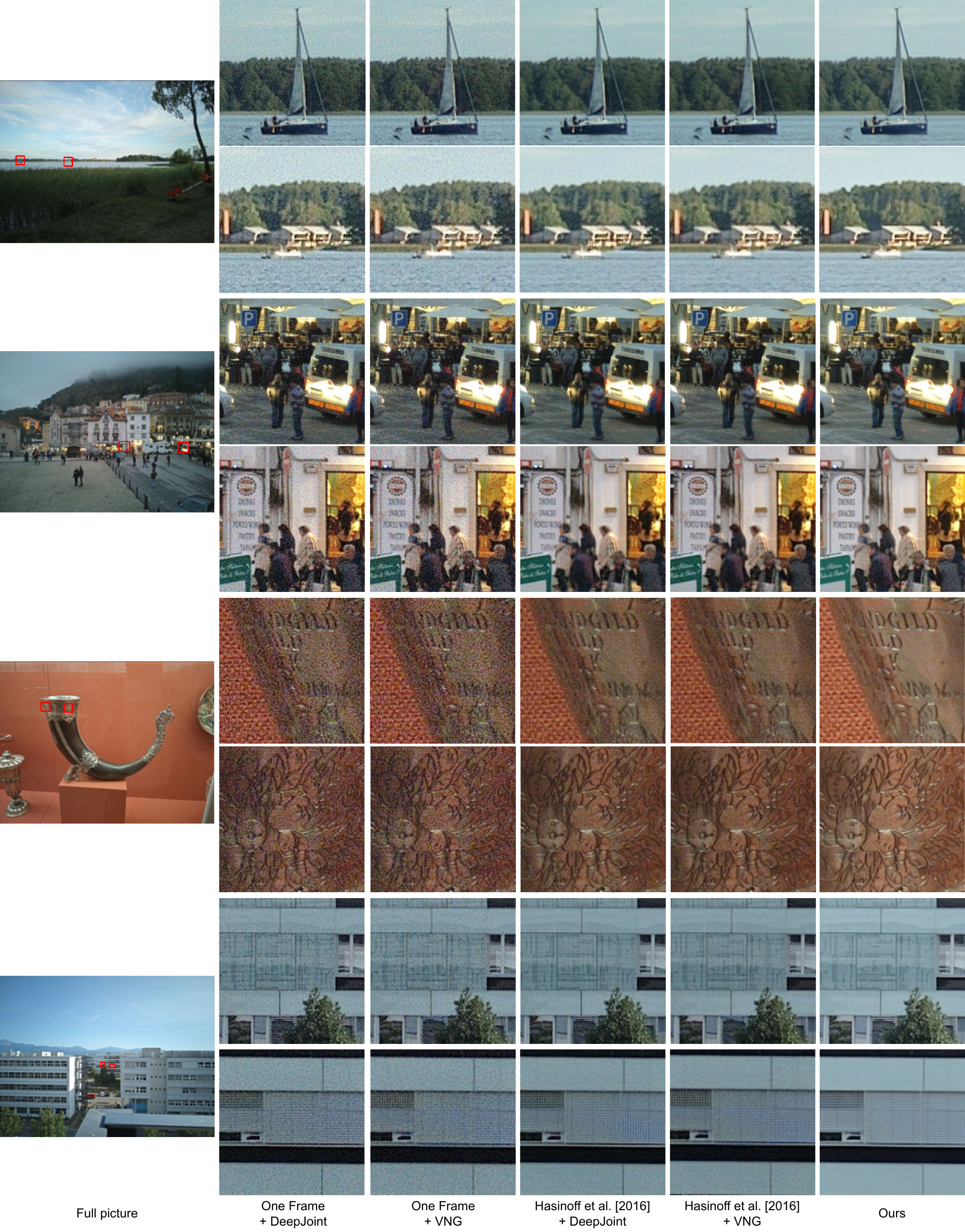}
  \caption{\textbf{Addional comparison with demosaicing techniques:}
  Our method compared with dcraw's Variable Number of Gradients~\cite{Chang:1999:CFA} and \textit{DeepJoint}~\cite{Gharbi:2016:DJD}. Both demosaicing techniques are applied to either one frame from a burst or result of burst merging as described in Hasinoff \etal ~\shortcite{Hasinoff:2016:BPH}.
  Readers are encouraged to zoom in aggressively (300\% or more).}
  \label{fig:results_demosaicing_supplement}
\end{figure*}

}
{
}

\end{document}